\documentclass[accepted]{uai2026}

\usepackage[american]{babel}

\usepackage{natbib} 
\usepackage{mathtools} 
\usepackage{booktabs} 
\usepackage{tikz} 

\usepackage{amsthm}
\newtheorem{lem}{Lemma}
\theoremstyle{theorem} \newtheorem{defn}{Definition}
\theoremstyle{definition} \newtheorem{assumption}{Assumption}
\theoremstyle{definition} 
\theoremstyle{theorem} \newtheorem{theorem}{Theorem}
\theoremstyle{corollary} \newtheorem{corollary}{Corollary}

\usepackage{amsmath}
\usepackage{blkarray}

\usepackage[ruled, vlined, linesnumbered]{algorithm2e}
\usepackage{lmodern}
\usepackage{amsfonts}
\SetKwProg{Init}{init}{}{}

\newcommand\reallywidehat[1]{%
\savestack{\tmpbox}{\stretchto{%
  \scaleto{%
    \scalerel*[\widthof{\ensuremath{#1}}]{\kern-.6pt\bigwedge\kern-.6pt}%
    {\rule[-\textheight/2]{1ex}{\textheight}}
  }{\textheight}%
}{0.5ex}}%
\stackon[1pt]{#1}{\tmpbox}%
}
\def\rightarrowCirc{\hbox{$\circ$}\kern-1.5pt\hbox{$\rightarrow$}}
\def\circHyphenCirc{\hbox{$\circ$}\kern-1.5pt\hbox{$-$}\kern-1.5pt\hbox{$\circ$}}
\def\circHyphen{\hbox{$\circ$}\kern-1.5pt\hbox{$-$}}

\def\rightarrowCirc{\hbox{$\circ$}\kern-1.5pt\hbox{$\rightarrow$}}
\def\circHyphenCirc{\hbox{$\circ$}\kern-1.5pt\hbox{$-$}\kern-1.5pt\hbox{$\circ$}}
\def\circHyphen{\hbox{$\circ$}\kern-1.5pt\hbox{$-$}}

\usepackage{array}
\usepackage{xcolor}
\usepackage{csvsimple}
\usepackage{multicol}

\usepackage{tabularx}
\usepackage{multirow}

\usepackage{hyperref}       
\usepackage{url}            

\title{Density Ratio-based Causal Discovery \\from Bivariate Continuous-Discrete Data}

\author[1,3,4]{Takashi Nicholas Maeda}
\author[2,3,4]{Shohei Shimizu}
\author[3]{Hidetoshi Matsui}
\affil[1]{%
    Gakushuin University, Japan
}
\affil[2]{%
    The University of Osaka, Japan
}
\affil[3]{%
    Shiga University, Japan
  }
\affil[4]{%
    RIKEN, Japan
  }
  
\begin{document}
\maketitle

\begin{abstract}
We address the problem of inferring the causal direction between a continuous variable $X$ and a discrete variable $Y$ from observational data. For the model $X \to Y$, we adopt the threshold model used in prior work. For the model $Y \to X$, we consider two cases: (1) the conditional distributions of $X$ given different values of $Y$ form a location-shift family, and (2) they are mixtures of generalized normal distributions with independently parameterized components. We establish identifiability of the causal direction through three theoretical results. First, we prove that under $X \to Y$, the density ratio of $X$ conditioned on different values of $Y$ is monotonic. Second, we establish that under $Y \to X$ with non-location-shift conditionals, monotonicity of the density ratio holds only on a set of Lebesgue measure zero in the parameter space. Third, we show that under $X \to Y$, the conditional distributions forming a location-shift family requires a precise coordination between the causal mechanism and input distribution, which is non-generic under the principle of independent mechanisms. Together, these results imply that monotonicity of the density ratio characterizes the direction $X \to Y$, whereas non-monotonicity or location-shift conditionals characterizes $Y \to X$. Based on this, we propose Density Ratio-based Causal Discovery (DRCD), a method that determines causal direction by testing for location-shift conditionals and monotonicity of the estimated density ratio. Experiments on synthetic and real-world datasets demonstrate that DRCD outperforms existing methods.
\end{abstract}

\begin{figure*}[t]
\centering
\includegraphics[width=14.0cm]{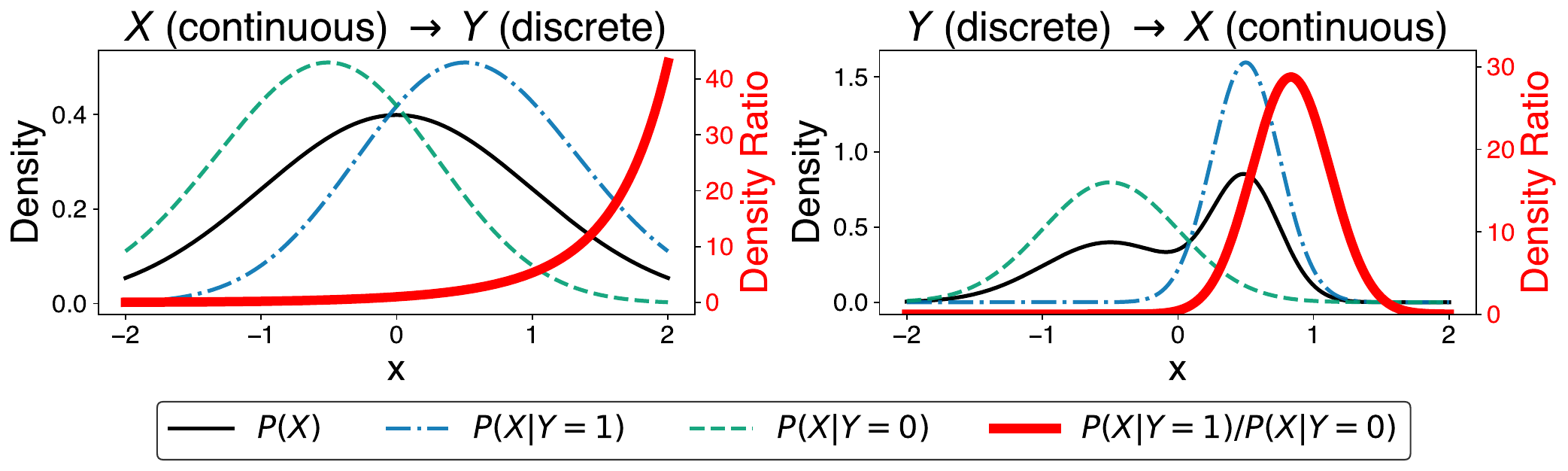}
\caption{Monotonicity property of the density ratio. When a continuous variable ($X$) causes a discrete variable ($Y$) (left), the conditional density ratio $\frac{P(X|Y=1)}{P(X|Y=0)}$ (red line) exhibits monotonic behavior. In contrast, when the causal direction is reversed ($Y \to X$) with non-location-shift conditionals (right), this monotonicity property does not hold. See Appendix~\ref{sec:example1} for the generative model settings used in this figure.}
\label{figure:plot}
\end{figure*}

\section{Introduction}
\label{sec:intro}

Determining the causal relationship between continuous and discrete variables from observational data alone is a fundamental problem in causal discovery.
Consider, for example, the relationship between a biological marker (a continuous variable) and the presence of a disease (a discrete variable).
It is often unclear whether changes in the biomarker cause the disease or whether early-stage disease processes alter the biomarker's levels.
While randomized controlled trials can resolve such ambiguities, they are frequently infeasible due to ethical, logistical, or financial constraints.

\textit{Causal discovery}~\citep{pearl2000, spirtes2000, 10.3389/fgene.2019.00524} aims to infer causal relationships from observational data without experiments.
Developing causal discovery methods requires three key steps: specifying assumptions about the data-generating process (the \textit{causal model}), proving that causal relationships are identifiable under these assumptions, and designing algorithms to estimate causal directions based on this identifiability.

Existing approaches for mixed continuous-discrete data face fundamental limitations.
Constraint-based methods~\citep{Spirtes91, fci} rely on conditional independence tests, which require additional variables for conditioning and thus cannot determine causal direction in bivariate settings.
Functional causal model-based approaches~\citep{ijcai2018p711, pmlr-v177-zeng22a, pmlr-v258-yao25a} assume that when a discrete variable $Y$ causes a continuous variable $X$, the conditional distributions $P(X|Y=c)$ form a location-shift family, differing only in their means while sharing the same shape and scale.
This assumption excludes causal models where conditional distributions have different shapes or variances.
Flexible score-based methods~\citep{Marx2019, GSF} avoid such distributional assumptions but face the challenge of fairly comparing causal directions between variables of inherently different types, as continuous and discrete variables differ fundamentally in their information content and scale.

For the model $X \to Y$, we adopt the threshold model from prior work~\citep{ijcai2018p711, pmlr-v177-zeng22a}.
For the model $Y \to X$, we extend existing frameworks by considering two cases: (1) the conditional distributions of $X$ given different values of $Y$ form a location-shift family, as assumed by prior methods~\citep{ijcai2018p711, pmlr-v177-zeng22a, pmlr-v258-yao25a}, and (2) they are mixtures of generalized normal distributions with independently specified components, allowing heterogeneous variances or shapes.

Our theoretical contributions are as follows:
\begin{itemize}
    \item We prove that under the model $X \to Y$, the density ratio $\frac{P(X|Y=c_t)}{P(X|Y=c_s)}$ is monotonic (Section~\ref{sec:idenfiability}).
    \item We show that under $Y \to X$ with non-location-shift conditionals (i.e., conditionals with different shapes or scales), monotonicity of the density ratio holds only on a set of Lebesgue measure zero in the parameter space (Section~\ref{sec:assumptions-non-generic}).
    \item We show that under $X \to Y$, the conditional distributions forming a location-shift family requires a precise coordination between the causal mechanism and input distribution, which is non-generic under the principle of independent mechanisms~\citep{PetJanSch17, 10.5555/3042573.3042635} (Section~\ref{sec:assumptions-non-generic}).
\end{itemize}
These results imply that monotonicity of the density ratio characterizes the direction $X \to Y$, whereas non-monotonicity or location-shift conditionals characterizes $Y \to X$. Figure~\ref{figure:plot} illustrates the monotonicity of the density ratio under $X \to Y$ and its absence under $Y \to X$ with non-location-shift conditionals.

Based on these theoretical findings, we propose \textit{Density Ratio-based Causal Discovery} (DRCD), which determines causal direction by testing for location-shift conditionals and monotonicity of the estimated density ratio (Section~\ref{sec:method}).
Since DRCD tests a property of the density ratio rather than comparing scores across causal directions, it avoids the difficulty of fairly comparing models between different variable types.
Experiments on synthetic and real-world datasets demonstrate that DRCD consistently outperforms existing methods (Section~\ref{sec:experiments}).

For our theoretical results, proofs are provided in Appendix~\ref{sec:proofs}. Code is available at \url{https://github.com/causal111/DRCD}.

\section{Related Studies}
\label{sec:related}

\textbf{Constraint-based approaches.}
Constraint-based methods such as PC~\citep{Spirtes91} and FCI~\citep{fci} determine causal structure through conditional independence tests.
While extensions to mixed data exist~\citep{Tsagris:2018aa, copulapc, 10.1093/nar/gkaa350}, these methods fundamentally cannot identify causal direction in bivariate settings, as there are no additional variables available for conditioning.

\textbf{Functional causal model-based approaches.}
Methods such as MIC~\citep{ijcai2018p711}, LiM~\citep{pmlr-v177-zeng22a}, and MANMs~\citep{pmlr-v258-yao25a} assume specific functional forms for the causal mechanisms.
For the direction $X \to Y$, they employ a threshold model where $Y$ is determined by whether $f(X) + N_Y$ exceeds a threshold.
For the direction $Y \to X$, they assume that the conditional distributions $P(X|Y=c)$ form a location-shift family, differing only in their means while sharing the same shape and scale.
This assumption excludes causal models with heterogeneous conditional distributions.
Our work extends this framework by additionally allowing non-location-shift conditionals under $Y \to X$.

\textbf{Flexible score-based methods.}
Methods such as CRACK~\citep{Marx2019} and GSF~\citep{GSF} avoid strong distributional assumptions by using flexible encoding schemes or kernel-based scores.
However, these methods face a fundamental difficulty in fairly comparing causal directions between continuous and discrete variables, which differ inherently in information content and scale.
Ad hoc normalization techniques have been proposed to address this issue, but they lack theoretical justification.
Our approach sidesteps this challenge entirely: rather than comparing scores across causal directions, it tests whether the density ratio is monotonic, eliminating the need for cross-type score comparison and ad hoc normalization.

\section{Preliminaries}
\label{sec:preliminaries}

We introduce key definitions used throughout this paper.

\begin{defn}[\textbf{Density ratio}]\label{defn:density-ratio}
For any two distinct values $c_s$ and $c_t$ of $Y$, the \textbf{density ratio} is defined as
\begin{equation}\label{eq:density-ratio}
G_{c_s,c_t}(x)=\frac{p_{X\mid Y}(x \mid c_t)}{p_{X\mid Y}(x \mid c_s)}.
\end{equation}
\end{defn}

\begin{defn}[\textbf{Monotonicity}]\label{defn:monotonic}
Throughout this paper, we use the term \textbf{monotonic} to refer exclusively to functions that are non-decreasing or non-increasing but not constant.
Specifically, a function $f$ is called monotonic if either (1) for all $x_0 < x_1$, $f(x_0) \leq f(x_1)$ and $f(-\infty) < f(\infty)$, or (2) for all $x_0 < x_1$, $f(x_0) \geq f(x_1)$ and $f(-\infty) > f(\infty)$.
\end{defn}

\begin{defn}[\textbf{Non-monotonicity}]\label{defn:non-monotonic}
A function $f$ is called \textbf{non-monotonic} if it is neither monotonic nor constant. Specifically, a function $f$ is non-monotonic if and only if there exist points $x_0 < x_1 < x_2$ such that $(f(x_0) - f(x_1))(f(x_1) - f(x_2)) < 0$.
\end{defn}

\begin{defn}[\textbf{Location-shift family}]\label{defn:location-shift}
Two distributions $P_{X|Y}(x|c_s)$ and $P_{X|Y}(x|c_t)$ belong to the same \textbf{location-shift family} if there exists a constant $\mu \neq 0$ such that $p_{X|Y}(x|c_t) = p_{X|Y}(x - \mu|c_s)$ for all $x$.
\end{defn}

\section{Model}
\label{sec:model}

For a continuous variable $X$ and a discrete variable $Y$, we consider three possible causal models: (1) $X$ causes $Y$, (2) $Y$ causes $X$, or (3) no causal relationship exists.
Throughout, we assume that there are no latent common causes of $X$ and $Y$. That is, all three models assume that the exogenous variables driving $X$ and $Y$ are determined by independent mechanisms.

\subsection{The model where $X$ is the cause of $Y$}
\label{sec:causal-model}

We consider a causal model where random variables $X$ and $Y$ satisfy
\begin{equation}
X = N_X, \ \ 
Z_i = \mathbb{I}[f_i(X) + N_{Y,i} \geq 0],\ \  
Y = q\bigl((Z_i)_{i=1}^m\bigr),
\end{equation}
where $N_X$ is a continuous random variable, $N_{Y,i}$ are continuous noise variables independent of $N_X$, and $m \in \mathbb{N}$ is the number of binary indicators $Z_i$. Let $\mathcal{S}_X$ denote the support of $X$. Each $f_i$ is a monotonic function defined on $\mathcal{S}_X$. The indicator function $\mathbb{I}[\cdot]$ equals 1 if its argument is true and 0 otherwise. The function $q$ maps the joint configuration of $(Z_i)_{i=1}^m$ to the discrete outcome $Y$.

We assume that the mechanism determining the distribution of $N_X$, the functions $f_i$, and the distributions of $N_{Y,i}$ are specified independently of each other. The implications of this assumption are detailed in Section~\ref{sec:assumptions-nonlocational}.

We additionally assume the following:

\begin{assumption}[\textbf{Separability of outcomes}]
\label{assumption:separability}
Distinct outcomes $c_s$ and $c_t$ of $Y$ are always distinguishable by at least one indicator $Z_i$:
\begin{equation}
c_s \neq c_t \;\Longrightarrow\; \exists i \in \{1,\dots,m\},\;\; Z_i \mid_{Y=c_s} \;\neq\; Z_i \mid_{Y=c_t}.
\end{equation}
Here $Z_i \mid_{Y=c_s}$ denotes the value of $Z_i$ for observations with outcome $Y=c_s$.
\end{assumption}

This ensures that different outcomes of $Y$ cannot coincide in all indicators $(Z_i)_{i=1}^m$.  
However, the converse does not necessarily hold: different configurations of $(Z_i)_{i=1}^m$ may map to the same value of $Y$.  

In other words, each $Z_i$ can be regarded as a noisy binary split of $X$, and $q$ assigns $Y$ based on the joint configuration of these splits.

\subsection{The model where $Y$ is the cause of $X$}
\label{sec:model-YX}

The causal model in this case is given by
\begin{align}
\label{eq:model-YX}
X = \sum_{k=1}^{K} \mathbb{I}[Y=c_k] N_{X,k}, \quad Y = N_Y,
\end{align}
where $N_Y$ is a discrete-valued random variable with $P(N_Y=c_k)=p_k$ such that $\sum_{k}p_k=1$, and each $N_{X,k}$ is a continuous random variable.

We consider two cases for the relationship among the conditional distributions $\{N_{X,k}\}_{k=1}^K$:

\paragraph{Case 1: Location-shift conditionals.}
The distributions of $\{N_{X,k}\}_{k=1}^K$ belong to the same location-shift family, i.e., they are identical up to a shift in location. In this case, there exist constants $\mu_k$ such that $p(N_{X,k}=x) = p(N_{X,1}=x-\mu_k)$ for all $x$ and $k$.

\paragraph{Case 2: Independently parameterized conditionals.}
The parameters governing each $N_{X,k}$ are independently parameterized, and the distributions do not belong to the same location-shift family. Specifically, each $N_{X,k}$ follows a finite mixture of generalized normal distributions:
\begin{align}
\label{eq:mixture-gen-normal}
p(N_{X,k}=x) = \sum_{j=1}^{L_k} \omega_{k,j} \, \varphi(x;\, \mu_{k,j}, \alpha_{k,j}, \beta_{k,j}),
\end{align}
where $\varphi(x;\, \mu, \alpha, \beta) = \frac{\beta}{2\alpha\Gamma(1/\beta)} \exp\!\left(-\left|\frac{x-\mu}{\alpha}\right|^\beta\right)$ is the generalized normal density with location $\mu \in \mathbb{R}$, scale $\alpha > 0$, and shape $\beta > 0$, $L_k \ge 1$ is the number of mixture components for value $c_k$, and $\omega_{k,j} \geq 0$ are the mixture weights satisfying $\sum_{j=1}^{L_k} \omega_{k,j} = 1$. All parameters are determined independently for each value of $Y$.
This reflects the causal intuition that each value of $Y$ independently determines the mechanism generating $X$. While related in spirit to the principle of independent mechanisms~\citep{PetJanSch17,10.5555/3042573.3042635}, which concerns the independence between the cause distribution and the causal mechanism, the independence here refers to the autonomous parameterization of conditional distributions across different values of the cause.

The mixture structure allows modeling of multimodal and asymmetric conditional distributions, and the generalized normal components provide flexibility in modeling both light-tailed and heavy-tailed distributions through the shape parameters $\beta_{k,j}$.
This class is closed under finite mixtures, so the model class in Equation~\eqref{eq:mixture-gen-normal} encompasses hierarchical mixture structures.

\subsection{The model with no causal relationship}
\label{sec:model-nocausal}

In the case where there is no causal relationship between $X$ and $Y$, the model is given by
\begin{align}
\label{eq:model-nocausal}
X = N_X, \quad Y = N_Y,
\end{align}
where $N_X$ is a continuous random variable and $N_Y$ is a discrete-valued random variable. We assume that $N_X$ and $N_Y$ are independent.

\paragraph{Examples in real-world scenarios.}
Our models capture causal structures that appear in various domains. In particular, the non-location-shift conditionals in Case~2, which go beyond the location-shift assumption of prior work, arise naturally in practice. Consider a causal chain $X_1 \to Y \to X_2$ where $X_1$ and $X_2$ are continuous and $Y$ is discrete. The link $X_1 \to Y$ corresponds to our threshold model (Section~\ref{sec:causal-model}), and the link $Y \to X_2$ corresponds to our heteroscedastic model where $X_2$ has distinct variances conditioned on $Y$ (Case~2 in Section~\ref{sec:model-YX}).
Such structures arise in economics~\citep{RePEc:ucp:jpolec:v:87:y:1979:i:5:p:s7-36}, where ability and family background ($X_1$) influence college attendance ($Y$), which in turn affects future income ($X_2$). In health economics~\citep{Manning:1987aa}, health status and demographics ($X_1$) determine the type of medical service usage ($Y$), which then influences medical expenditure ($X_2$). In developmental biology~\citep{doi:10.1126/sciadv.adf3497}, cell size at birth ($X_1$) triggers meristemoid behavior ($Y$), affecting the number of cell divisions ($X_2$).

\section{Genericity Analysis}
\label{sec:assumptions-non-generic}

In this section, we show that certain configurations in our models are non-generic, and analyze the conditions under which they arise. These results play a key role in establishing identifiability in Section~\ref{sec:idenfiability}.

\subsection{Non-location-shift conditionals under $X \to Y$}
\label{sec:assumptions-nonlocational}

Under the model $X \to Y$ described in Section~\ref{sec:causal-model}, we show that the conditional distributions $P(X|Y=c_s)$ and $P(X|Y=c_t)$ generically do not form a location-shift family.

\begin{lem}[\textbf{Location-shift requires special structure}]
\label{lem:location-shift-rare}
Assume that $(X,Y)$ are generated according to the model in Section~\ref{sec:causal-model} with causal direction $X \to Y$. Let $c_s$ and $c_t$ be distinct outcomes of $Y$, and let $i$ be an index such that $Z_i$ separates $c_s$ and $c_t$ as guaranteed by Assumption~\ref{assumption:separability}. Let $F_i(\cdot)$ denote the cumulative distribution function (CDF) of the noise variable $N_{Y,i}$. If the conditional distributions $P(X|Y=c_s)$ and $P(X|Y=c_t)$ belong to the same location-shift family with shift $\mu \neq 0$, then the mechanism function $f_i$, the input density $p_X$, and the noise CDF $F_i$ must satisfy
\begin{align}
\label{eq:location-shift-condition}
\frac{1 - F_i(-f_i(x))}{F_i(-f_i(x - \mu))} = C \cdot \frac{p_X(x - \mu)}{p_X(x)}
\end{align}
for all $x$, where $C > 0$ is a constant determined by the marginal probabilities of $Y$.
\end{lem}

The left-hand side of Equation~\eqref{eq:location-shift-condition} is determined solely by the causal mechanism (characterized by $f_i$ and the noise CDF $F_i$), while the right-hand side depends exclusively on the marginal density of the cause, $p_X$.
According to the principle of independent mechanisms~\citep{PetJanSch17,10.5555/3042573.3042635}, the mechanism generating the effect is autonomous and contains no information about the distribution of the cause.
Consequently, it is non-generic for the mechanism and the input distribution to be coordinated in such a way that Equation~\eqref{eq:location-shift-condition} holds for all $x$. We therefore state the non-location-shift property under $X \to Y$ as an explicit genericity condition:

\begin{assumption}[\textbf{Genericity condition: non-location-shift conditionals under $X \to Y$}]
\label{assumption:non-location-shift}
Under the model $X \to Y$, for any distinct outcomes $c_s$ and $c_t$, the conditional distributions $P(X|Y=c_s)$ and $P(X|Y=c_t)$ do not belong to the same location-shift family.
\end{assumption}

\subsection{Non-monotonicity of density ratio of conditionals under $Y \to X$}
\label{sec:assumptions-tail}

Under the model $Y \to X$ described in Section~\ref{sec:model-YX}, when the conditional distributions $P(X|Y=c_s)$ and $P(X|Y=c_t)$ do not belong to the same location-shift family (Case~2), we argue that the density ratio $G_{c_s,c_t}(x)$ is generically non-monotonic. We focus on Case~2 here, as Case~1 (location-shift conditionals) is identified not through monotonicity of the density ratio but through whether the conditionals form a location-shift family.

\begin{lem}[\textbf{Non-generic monotonicity of density ratios}]
\label{lem:monotonicity-conditions}
Let $f$ and $g$ be finite mixtures of generalized normal distributions:
\begin{align*}
f(x) &= \sum_{j=1}^{L_f} \omega^f_j \, \varphi(x;\, \mu^f_j, \alpha^f_j, \beta^f_j), \\
g(x) &= \sum_{j=1}^{L_g} \omega^g_j \, \varphi(x;\, \mu^g_j, \alpha^g_j, \beta^g_j),
\end{align*}
where $\varphi(x;\, \mu, \alpha, \beta) = \frac{\beta}{2\alpha\Gamma(1/\beta)} \exp\!\left(-\left|\frac{x-\mu}{\alpha}\right|^\beta\right)$ is the generalized normal density with location $\mu \in \mathbb{R}$, scale $\alpha > 0$, and shape $\beta > 0$. We parameterize the mixture weights by their first $L_f-1$ (resp.\ $L_g-1$) components, with $\omega^f_{L_f} = 1 - \sum_{j=1}^{L_f-1}\omega^f_j$ (resp.\ $\omega^g_{L_g} = 1 - \sum_{j=1}^{L_g-1}\omega^g_j$), subject to all weights being strictly positive. Write $\theta_f = (\omega^f_{1:L_f-1},\, \mu^f,\, \alpha^f,\, \beta^f)$ and $\theta_g = (\omega^g_{1:L_g-1},\, \mu^g,\, \alpha^g,\, \beta^g)$ for the parameter vectors of $f$ and $g$, respectively. Define the following two parameter spaces:
\begin{itemize}
\item[\textup{(a)}] $\Theta_a$: all entries of $(\theta_f, \theta_g)$ are free, so that $\Theta_a$ is an open subset of $\mathbb{R}^{4L_f+4L_g-2}$.
\item[\textup{(b)}] $\Theta_b$: the shape parameters $\beta^f$ and $\beta^g$ are fixed to arbitrary positive constants and all remaining entries of $(\theta_f, \theta_g)$ are free, so that $\Theta_b$ is an open subset of $\mathbb{R}^{3L_f+3L_g-2}$.
\end{itemize}
Let $\mathcal{M}_a = \{\theta \in \Theta_a : r(x) = f(x)/g(x) \text{ is monotonic}\}$ and $\mathcal{M}_b = \{\theta \in \Theta_b : r(x) = f(x)/g(x) \text{ is monotonic}\}$. Then $\mathcal{M}_a$ has Lebesgue measure zero in $\Theta_a$, and $\mathcal{M}_b$ has Lebesgue measure zero in $\Theta_b$.
\end{lem}

Case~(a) corresponds directly to the independently parameterized conditionals in our model (Case~2 in Section~\ref{sec:model-YX}), where all parameters of each conditional distribution are free. Case~(b) covers the important special case where the component family is fixed a priori, such as Gaussian mixtures ($\beta^f_j = \beta^g_j = 2$) or Laplace mixtures ($\beta^f_j = \beta^g_j = 1$), while the remaining parameters (mixture weights, locations, and scales) are free. By treating these two cases separately, Lemma~\ref{lem:monotonicity-conditions} ensures that non-genericity of monotonicity holds both in the fully flexible setting assumed by our model and in restricted-family settings commonly encountered in practice.

\begin{corollary}
\label{cor:nonmonotonic}
Under the model $Y \to X$ with non-location-shift conditionals (Case~2 in Section~\ref{sec:model-YX}), the density ratio $G_{c_s,c_t}(x)$ is monotonic only on a set of Lebesgue measure zero in the parameter space.
\end{corollary}

\section{Identifiability}
\label{sec:idenfiability}

This section establishes that the causal relationship between a continuous variable $X$ and a discrete variable $Y$ is identifiable under the modeling assumptions in Section~\ref{sec:model} and the genericity analysis in Section~\ref{sec:assumptions-non-generic}.

\subsection{Properties under $X \to Y$}

By Lemma~\ref{lem:location-shift-rare} and Assumption~\ref{assumption:non-location-shift} in Section~\ref{sec:assumptions-nonlocational}, the conditional distributions $P(X|Y=c_s)$ and $P(X|Y=c_t)$ do not belong to the same location-shift family when $X \to Y$. Furthermore, the density ratio $G_{c_s,c_t}(x)$ is monotonic:

\begin{lem}[\textbf{Monotonicity of the density ratio under $X \to Y$}]
\label{lem:main}
Assume that $(X,Y)$ are generated according to the model described in Section~\ref{sec:causal-model} with the causal direction $X \to Y$. Then, for any distinct outcomes $c_s$ and $c_t$ of $Y$, the density ratio $G_{c_s,c_t}(x)$ is monotonic in $x$.
\end{lem}

\subsection{Properties under $Y \to X$}

Under the model $Y \to X$ described in Section~\ref{sec:model-YX}, the conditional distributions $P(X|Y=c_s)$ and $P(X|Y=c_t)$ either belong to the same location-shift family or do not. In the latter case, by Corollary~\ref{cor:nonmonotonic}, the density ratio $G_{c_s,c_t}(x)$ is non-monotonic except on a set of Lebesgue measure zero in the parameter space.

\subsection{Properties under no causal relationship}

\begin{lem}[\textbf{Identical conditionals under no causal relationship}]
\label{lem:no_causal}
The conditional distributions $P_{X|Y}(x|c_s)$ and $P_{X|Y}(x|c_t)$ are identical for all distinct $c_s$ and $c_t$ if and only if $X$ and $Y$ have no causal relationship. Equivalently, $G_{c_s,c_t}(x) = 1$ for all $x$ if and only if $X$ and $Y$ are independent.
\end{lem}

\subsection{Identifiability theorem}

The properties established in the preceding subsections lead to the following identifiability result.

\begin{theorem}[\textbf{Identifiability of the causal relationship}]
\label{thm:identifiability}
Assume that the joint distribution of $(X,Y)$ is generated according to exactly one of the three causal models defined in Section~\ref{sec:model}: (1) $X \to Y$, (2) $Y \to X$, or (3) no causal relationship. Then the causal relationship between $X$ and $Y$ is identifiable.
\end{theorem}

\section{Method}
\label{sec:method}
We propose \textit{Density Ratio-based Causal Discovery} (DRCD), a method for determining which of the three causal models best characterizes the relationship between a continuous variable $X$ and a discrete variable $Y$: (1) $X$ causes $Y$, (2) $Y$ causes $X$, or (3) no causal relationship exists between $X$ and $Y$. DRCD consists of four main steps: (1) testing for causal existence, (2) testing for location-shift relationship, (3) density ratio estimation, and (4) monotonicity evaluation for causal direction determination. The DRCD procedure is formalized in Algorithm~\ref{algo:drcd}; its computational complexity is analyzed in Appendix~\ref{sec:complexity}. Below, we provide a detailed explanation of each step and clarify how our procedure operationalizes the theoretical properties established in Sections~\ref{sec:assumptions-non-generic} and~\ref{sec:idenfiability}.

\begin{algorithm}[h]
\SetKwProg{init}{initialization}{}{}
\DontPrintSemicolon
\KwIn{$D$: Dataset of continuous $X$ and discrete $Y$, $\alpha$: Significance level, $\gamma$: Location-shift threshold, $\rho$: Monotonicity threshold, $n_{\rm points}$: Number of grid points.}
\KwOut{Estimated causal relationship}
\SetKwBlock{Begin}{function}{end function}
\Begin($\text{DRCD} {(} D, \alpha, \gamma, \rho, n_{\rm points} {)}$)
{
$c_s, c_t\leftarrow$ the two distinct values of $Y$ with the largest frequencies\;
$X_{c_s} \leftarrow \{x \in X \mid Y=c_s\}$\; $X_{c_t} \leftarrow \{x \in X \mid Y=c_t\}$\;
\textsf{// Step 1: Test for causal existence.}\;
\If{KS-test$(X_{c_s}, X_{c_t})$ $> \alpha$}{
    \Return{``No causal relationship''}}
\textsf{// Step 2: Test for location-shift.}\;
$\tilde{X}_{c_s} \leftarrow X_{c_s} - \text{mean}(X_{c_s})$\; $\tilde{X}_{c_t} \leftarrow X_{c_t} - \text{mean}(X_{c_t})$\;
\If{KS-test$(\tilde{X}_{c_s}, \tilde{X}_{c_t})$ $> \gamma$}{
    \Return{``$Y \to X$''}}
\textsf{// Step 3: Estimate density ratio.}\;
$G_{c_s,c_t}(x)\leftarrow$ uLSIF estimate of $\frac{p_{X|Y}(x|c_t)}{p_{X|Y}(x|c_s)}$\;
$[x_{\rm min}, x_{\rm max}] \leftarrow \text{supp}(X_{c_s}) \cap \text{supp}(X_{c_t})$\;
\textsf{// Step 4: Evaluate monotonicity.}\;
$\mathcal{X} \leftarrow$ $n_{\rm points}$ equidistant points in $[x_{\rm min}, x_{\rm max}]$\;
$r_s \leftarrow$ Spearman correlation between indices and $G_{c_s,c_t}(\mathcal{X})$\;
\If{$|r_s| > \rho$}{
    \Return{``$X \to Y$''}}
\Else{\Return{``$Y \to X$''}}
}
\caption{The DRCD algorithm}
\label{algo:drcd}
\end{algorithm}

\textbf{Step 1: Testing for causal existence.}  
To assess the existence of a causal relationship between $X$ and $Y$, we examine whether the conditional distributions $p_{X|Y}(x|y)$ differ across values of $y$. When no causal relationship exists, $X$ and $Y$ are statistically independent, so these distributions should be identical. Conversely, under the models $X \to Y$ and $Y \to X$ defined in Sections~\ref{sec:causal-model} and~\ref{sec:model-YX}, the conditional distributions $p_{X|Y}(x|c_s)$ and $p_{X|Y}(x|c_t)$ differ for any distinct values $c_s$ and $c_t$ (by Assumption~\ref{assumption:separability} and the model definitions). Therefore, it suffices to test a single pair. We select the two most frequent values of $Y$ as $c_s$ and $c_t$ to improve statistical reliability, and let $X_{c_s}$ and $X_{c_t}$ denote the corresponding subsamples of $X$. We apply the two-sample Kolmogorov--Smirnov (KS) test to $X_{c_s}$ and $X_{c_t}$. If no significant difference is detected, we conclude that there is no causal relationship; otherwise, we proceed to determine the causal direction.

\textbf{Step 2: Testing for location-shift relationship.}
By Assumption~\ref{assumption:non-location-shift}, a location-shift relationship between the conditionals implies $Y \to X$. We test this by centering each conditional sample (subtracting its mean) and applying the KS test to the centered samples. If the p-value exceeds the threshold $\gamma$, we infer $Y \to X$; otherwise, we proceed to Steps~3 and~4.

\textbf{Step 3: Density ratio estimation.}  
Having ruled out the location-shift case, we now estimate the density ratio to distinguish between $X \to Y$ and $Y \to X$ (non-location-shift). Based on Corollary~\ref{cor:nonmonotonic} and Lemma~\ref{lem:main}, the monotonicity of the density ratio generically determines the causal direction. We estimate the density ratio $G_{c_s,c_t}(x)=\frac{p_{X|Y}(x|c_t)}{p_{X|Y}(x|c_s)}$ using uLSIF~\citep{JMLR:v10:kanamori09a}. To ensure reliable estimation, we define a valid domain $[x_{\rm min}, x_{\rm max}]$ for evaluation, where $x_{\rm min} = \max\{\min\{X \mid Y = c_s\}, \min\{X \mid Y = c_t\}\}$ and $x_{\rm max} = \min\{\max\{X \mid Y = c_s\}, \max\{X \mid Y = c_t\}\}$. This restriction ensures that the evaluation region lies within the overlapping support of the two conditional distributions.

\begin{table*}[h]
\centering
\caption{Accuracy (\%) of causal discovery methods on synthetic datasets with 95\% confidence intervals. Accuracies above 80\% are shown in \textbf{bold}.}
\begin{tabular}{lcccc}
\toprule
Method & No Causation & $X \to Y$ & \multicolumn{2}{c}{$Y \to X$} \\
\cmidrule(lr){4-5}
 & & & Location-shift & Non-location-shift \\
\midrule
DRCD & \textbf{94.7\%} {\footnotesize(93.3--96.0\%)} & \textbf{91.8\%} {\footnotesize(90.1--93.5\%)} & \textbf{84.9\%} {\footnotesize(82.6--87.1\%)} & \textbf{85.4\%} {\footnotesize(83.2--87.6\%)} \\
LiM & 3.3\% {\footnotesize(2.2--4.4\%)} & 61.1\% {\footnotesize(58.1--64.1\%)} & \textbf{98.9\%} {\footnotesize(98.2--99.5\%)} & 68.1\% {\footnotesize(65.2--71.0\%)} \\
MIC & \textbf{86.1\%} {\footnotesize(83.9--88.2\%)} & 74.4\% {\footnotesize(71.7--77.1\%)} & 46.9\% {\footnotesize(43.8--50.0\%)} & 5.5\% {\footnotesize(4.1--7.0\%)} \\
MANMs & \textbf{95.4\%} {\footnotesize(94.1--96.7\%)} & \textbf{93.5\%} {\footnotesize(91.9--95.0\%)} & \textbf{96.5\%} {\footnotesize(95.3--97.6\%)} & 4.9\% {\footnotesize(3.6--6.3\%)} \\
CRACK & 1.9\% {\footnotesize(1.1--2.8\%)} & 0.0\% {\footnotesize(0.0--0.0\%)} & \textbf{99.6\%} {\footnotesize(99.2--99.9\%)} & \textbf{99.1\%} {\footnotesize(98.5--99.6\%)} \\
GSF & \textbf{100.0\%} {\footnotesize(100.0--100.0\%)} & 59.2\% {\footnotesize(56.1--62.2\%)} & \textbf{81.5\%} {\footnotesize(79.1--83.9\%)} & 0.4\% {\footnotesize(0.1--0.8\%)} \\
\bottomrule
\end{tabular}
\label{tab:accuracy_results}
\end{table*}

\textbf{Step 4: Evaluating monotonicity to determine causal direction.}  
We determine the causal direction by evaluating the monotonicity of the estimated density ratio $G_{c_s,c_t}(x)$. Within the evaluation region $[x_{\rm min}, x_{\rm max}]$, we construct an evenly spaced sequence $\mathcal{X} = (x_k)_{k=1}^{n_{\rm points}}$, where $n_{\rm points}$ denotes the number of grid points, and evaluate the density ratio at these points.
To assess monotonicity, we compute the Spearman rank correlation coefficient $r_s$ between the indices $(1, 2, \ldots, n_{\rm points})$ and the density ratio values $(G_{c_s,c_t}(x_1), G_{c_s,c_t}(x_2), \ldots, G_{c_s,c_t}(x_{n_{\rm points}}))$. If $|r_s| > \rho$, where $\rho$ is the monotonicity threshold, we conclude that the density ratio exhibits a strong monotonic pattern and infer $X \to Y$; otherwise, we infer $Y \to X$.

\noindent\textbf{Remark.}
Steps 1--4 of Algorithm~\ref{algo:drcd} operationalize the theoretical results as follows: Step~1 uses Lemma~\ref{lem:no_causal}; Step~2 uses Assumption~\ref{assumption:non-location-shift}; Steps~3--4 use Lemma~\ref{lem:main} and Corollary~\ref{cor:nonmonotonic}.

\begin{table*}[t]
\centering
\caption{Causal discovery results on the UCI Heart Disease dataset. Correct inferences are shown in \textbf{bold}. $X$ and $Y$ denote numerical and categorical variables, respectively.}
\label{tab:heart_results}
\begin{tabular}{llccccccc}
\toprule
$X$ & $Y$ & Ground truth & DRCD & LiM & MIC & MANMs & CRACK & GSF \\
\midrule
trestbps & sex & $Y \to X$ & No caus. & \textbf{Y$\to$X} & \textbf{Y$\to$X} & No caus. & \textbf{Y$\to$X} & No caus. \\
chol & sex & $Y \to X$ & \textbf{Y$\to$X} & X$\to$Y & \textbf{Y$\to$X} & X$\to$Y & \textbf{Y$\to$X} & X$\to$Y \\
age & num & $X \to Y$ & \textbf{X$\to$Y} & Y$\to$X & Y$\to$X & \textbf{X$\to$Y} & Y$\to$X & \textbf{X$\to$Y} \\
age & fbs & $X \to Y$ & \textbf{X$\to$Y} & Y$\to$X & Y$\to$X & Y$\to$X & \textbf{X$\to$Y} & No caus. \\
\midrule
\multicolumn{3}{l}{Correct inferences} & \textbf{3} & 1 & 2 & 1 & \textbf{3} & 1 \\
\multicolumn{3}{l}{Reversed inferences} & \textbf{0} & 3 & 2 & 2 & 1 & 1 \\
\bottomrule
\end{tabular}
\end{table*}

\begin{table*}[t]
\centering
\caption{Causal discovery results on the T\"ubingen Cause-Effect Pairs Dataset. Correct inferences are shown in \textbf{bold}. $X$ and $Y$ denote numerical and categorical variables, respectively.}
\label{tab:tubingen_combined_unified}
\begin{tabular}{lllccccccc}
\toprule
ID & $X$ & $Y$ & Ground truth & DRCD & LiM & MIC & MANMs & CRACK & GSF \\
\midrule
0005 & Shell length & Rings & $Y \to X$ & \textbf{Y$\to$X} & X$\to$Y & X$\to$Y & X$\to$Y & \textbf{Y$\to$X} & \textbf{Y$\to$X} \\
0094 & Temp. & Hour & $Y \to X$ & No caus. & X$\to$Y & X$\to$Y & X$\to$Y & \textbf{Y$\to$X} & No caus. \\
0095 & Elec. load & Hour & $Y \to X$ & \textbf{Y$\to$X} & X$\to$Y & \textbf{Y$\to$X} & X$\to$Y & \textbf{Y$\to$X} & X$\to$Y \\
0107 & Contrast & Answer & $X \to Y$ & \textbf{X$\to$Y} & Y$\to$X & \textbf{X$\to$Y} & \textbf{X$\to$Y} & Y$\to$X & \textbf{X$\to$Y} \\
\midrule
\multicolumn{4}{l}{Correct inferences} & \textbf{3} & 0 & 2 & 1 & \textbf{3} & 2 \\
\multicolumn{4}{l}{Reversed inferences} & \textbf{0} & 4 & 2 & 3 & 1 & 2 \\
\bottomrule
\end{tabular}
\end{table*}

\section{Experiments}
\label{sec:experiments}
In this section, we present comparative experiments using both synthetic and real-world datasets. The methods compared in our evaluation include LiM, MIC, MANMs, CRACK, and GSF. Detailed experimental settings are provided in Appendix~\ref{sec:expedetails}.

\subsection{Experiments with Synthetic Data}
\label{sec:artificialexperiments}
We conducted experiments using synthetic data generated under four causal scenarios. In all cases, we use independent noise variables $\xi_1$ to $\xi_9$, where $\xi_1, \xi_2, \xi_3, \xi_4, \xi_5, \xi_7, \xi_8$ are drawn from a mixture distribution $p(x) = 0.5 \cdot \mathcal{N}(x; \mu_1, \sigma_1^2) + 0.5 \cdot \text{Laplace}(x; \mu_2, b_2)$, with $\mu_1, \mu_2 \sim \mathcal{U}(-2, 2)$ and $\sigma_1, b_2 \sim \mathcal{U}(0.5, 2)$ independently sampled for each variable, $\xi_6$ follows the same mixture with fixed scale parameters ($\sigma = b = 1$), and $\xi_9$ follows the same mixture with scale parameters twice those of $\xi_8$.

\textbf{Case 1 (No causation):}
$X = \xi_1$ and $Y = \mathbb{I}[\xi_2 > 0]$.

\textbf{Case 2 ($X$ causes $Y$):}
$X = \xi_3$, with $Y = \mathbb{I}[aX + \xi_4 \geq 0]$, where $a \in \{-1, 1\}$ is chosen uniformly at random.

\textbf{Case 3 ($Y$ causes $X$, location-shift):}
$Y = \mathbb{I}[\xi_5 > 0]$, with $X = \mu \cdot Y + \xi_6$, where $\mu \in \{-2, 2\}$ is a fixed constant.

\textbf{Case 4 ($Y$ causes $X$, non-location-shift):}
$Y = \mathbb{I}[\xi_7 > 0]$, with $X = (1-Y)\xi_8 + Y\xi_9$.

For each case, we generated 1000 datasets, with each dataset containing 1000 observations, and conducted experiments on all datasets.

Table~\ref{tab:accuracy_results} presents the accuracy results. Values in parentheses indicate 95\% confidence intervals calculated using the binomial distribution. DRCD is the only method that consistently maintains high accuracy ($>80\%$) across all causal scenarios. Notably, LiM, MIC, and MANMs perform poorly in Case~4 (non-location-shift). This is expected because these methods assume that the conditional distributions $P(X|Y)$ form a location-shift family under $Y \to X$, an assumption violated in Case~4.
The hyperparameters of DRCD used here are $(\rho, \gamma) = (0.8, 0.5)$; a detailed sensitivity analysis examining how performance varies with these choices and the number of observations is provided in Appendix~\ref{sec:sensitivity}.

\subsection{Experiments using real-world data}
\label{sec:realexp}

We conducted comparative experiments using two real-world data sources: the UCI Heart Disease dataset~\citep{heart_disease_45} and the T\"ubingen cause-effect pairs dataset~\citep{JMLR:v17:14-518}. While real-world processes may not strictly follow the idealized assumptions of any causal discovery model, these experiments evaluate the robustness of each method in identifying causal signals within complex and noisy environments.

\paragraph{UCI Heart Disease Dataset.}
The UCI Heart Disease dataset contains numerical variables (\texttt{age}: patient age, \texttt{chol}: serum cholesterol, \texttt{trestbps}: resting blood pressure) and categorical variables (\texttt{sex}: gender, \texttt{fbs}: fasting blood sugar status, \texttt{num}: heart disease diagnosis).

We selected variable pairs for which the causal direction can be unambiguously determined from domain knowledge. Specifically, \texttt{sex} and \texttt{age} are inherent attributes that can only be causes, not effects. Since the dataset contains variables related to cardiovascular disease and diabetes, we referred to epidemiological studies from the Framingham Heart Study, a long-term cardiovascular cohort study, on cardiovascular disease risk factors~\citep{Pencina:2009aa} and on blood pressure and diabetes risk~\citep{Wei:2011aa}. We excluded variable pairs for which the ground truth causal direction could not be reliably established. Details are provided in Appendix~\ref{sec:causal_evidence}.

Table~\ref{tab:heart_results} shows the results. DRCD and CRACK achieved the highest accuracy (3 out of 4 correct). However, when DRCD made an error, it inferred ``No causation'' rather than the incorrect causal direction, whereas CRACK made one reversed inference.

\paragraph{T\"ubingen Cause-Effect Pairs Dataset.}
The T\"ubingen cause-effect pairs dataset is a widely used benchmark for bivariate causal discovery. From this dataset, we selected all pairs in which one variable is continuous and the other is discrete with a small number of distinct values. We excluded pairs where the discrete variable is a fine-grained discretization of an underlying continuous quantity with a natural linear ordering (e.g., age), as such variables are more appropriately treated as continuous rather than categorical. This criterion yielded four pairs, whose details are summarized in Table~\ref{tab:tubingen_combined_unified}.
As shown in the table, DRCD and CRACK achieved the highest accuracy, correctly identifying the causal direction for 3 out of 4 pairs. Notably, DRCD made no reversed inferences, whereas CRACK made one.

\section{Conclusion}
\label{sec:conclusion}
We proposed DRCD, a method for causal discovery from bivariate mixed data that determines causal direction by testing monotonicity of the conditional density ratio and detecting location-shift conditionals, avoiding both restrictive distributional assumptions and ad hoc normalization between variable types. The identifiability of DRCD rests on two complementary arguments: a measure-zero result for the non-location-shift case, and the principle of independent mechanisms for the location-shift case.

Our framework has limitations. We assume the absence of latent confounders and restrict attention to the bivariate setting. One promising direction is to integrate DRCD's monotonicity test as a local orientation criterion within constraint-based methods such as PC~\citep{Spirtes91} or FCI~\citep{fci}.
More broadly, testing monotonicity of the density ratio, rather than comparing scores across causal directions, may offer a principled alternative for causal discovery between variables of different types.

%

\bibliography{ref}

\begin{thebibliography}{23}
\providecommand{\natexlab}[1]{#1}
\providecommand{\url}[1]{\texttt{#1}}
\expandafter\ifx\csname urlstyle\endcsname\relax
  \providecommand{\doi}[1]{doi: #1}\else
  \providecommand{\doi}{doi: \begingroup \urlstyle{rm}\Url}\fi

\bibitem[Cui et~al.(2016)Cui, Groot, and Heskes]{copulapc}
Ruifei Cui, Perry Groot, and Tom Heskes.
\newblock Copula {PC} algorithm for causal discovery from mixed data.
\newblock In Paolo Frasconi, Niels Landwehr, Giuseppe Manco, and Jilles
  Vreeken, editors, \emph{Machine Learning and Knowledge Discovery in
  Databases}, pages 377--392, Cham, 2016. Springer International Publishing.
\newblock ISBN 978-3-319-46227-1.

\bibitem[Ge et~al.(2020)Ge, Raghu, Chrysanthis, and Benos]{10.1093/nar/gkaa350}
Xiaoyu Ge, Vineet~K Raghu, Panos~K Chrysanthis, and Panayiotis~V Benos.
\newblock {CausalMGM}: an interactive web-based causal discovery tool.
\newblock \emph{Nucleic Acids Research}, 48\penalty0 (W1):\penalty0 W597--W602,
  05 2020.
\newblock ISSN 0305-1048.
\newblock \doi{10.1093/nar/gkaa350}.
\newblock URL \url{https://doi.org/10.1093/nar/gkaa350}.

\bibitem[Glymour et~al.(2019)Glymour, Zhang, and
  Spirtes]{10.3389/fgene.2019.00524}
Clark Glymour, Kun Zhang, and Peter Spirtes.
\newblock Review of causal discovery methods based on graphical models.
\newblock \emph{Frontiers in Genetics}, Volume 10 - 2019, 2019.
\newblock ISSN 1664-8021.
\newblock \doi{10.3389/fgene.2019.00524}.
\newblock URL
  \url{https://www.frontiersin.org/journals/genetics/articles/10.3389/fgene.2019.00524}.

\bibitem[Gong et~al.(2023)Gong, Dale, Fung, Amador, Smit, and
  Bergmann]{doi:10.1126/sciadv.adf3497}
Yan Gong, Renee Dale, Hannah~F. Fung, Gabriel~O. Amador, Margot~E. Smit, and
  Dominique~C. Bergmann.
\newblock A cell size threshold triggers commitment to stomatal fate in
  {Arabidopsis}.
\newblock \emph{Science Advances}, 9\penalty0 (38), 2023.
\newblock \doi{10.1126/sciadv.adf3497}.
\newblock URL \url{https://www.science.org/doi/abs/10.1126/sciadv.adf3497}.

\bibitem[Huang et~al.(2018)Huang, Zhang, Lin, Sch\"{o}lkopf, and Glymour]{GSF}
Biwei Huang, Kun Zhang, Yizhu Lin, Bernhard Sch\"{o}lkopf, and Clark Glymour.
\newblock Generalized score functions for causal discovery.
\newblock In \emph{Proceedings of the 24th ACM SIGKDD International Conference
  on Knowledge Discovery \& Data Mining}, KDD '18, pages 1551--1560, New York,
  NY, USA, 2018. Association for Computing Machinery.
\newblock ISBN 9781450355520.
\newblock \doi{10.1145/3219819.3220104}.
\newblock URL \url{https://doi.org/10.1145/3219819.3220104}.

\bibitem[Janosi et~al.(1989)Janosi, Steinbrunn, Pfisterer, and
  Detrano]{heart_disease_45}
Andras Janosi, William Steinbrunn, Matthias Pfisterer, and Robert Detrano.
\newblock {Heart Disease}.
\newblock UCI Machine Learning Repository, 1989.
\newblock {DOI}: https://doi.org/10.24432/C52P4X.

\bibitem[Kanamori et~al.(2009)Kanamori, Hido, and
  Sugiyama]{JMLR:v10:kanamori09a}
Takafumi Kanamori, Shohei Hido, and Masashi Sugiyama.
\newblock A least-squares approach to direct importance estimation.
\newblock \emph{Journal of Machine Learning Research}, 10\penalty0
  (48):\penalty0 1391--1445, 2009.
\newblock URL \url{http://jmlr.org/papers/v10/kanamori09a.html}.

\bibitem[Manning et~al.(1987)Manning, Newhouse, Duan, Keeler, Leibowitz, and
  Marquis]{Manning:1987aa}
W~G Manning, J~P Newhouse, N~Duan, E~B Keeler, A~Leibowitz, and M~S Marquis.
\newblock Health insurance and the demand for medical care: evidence from a
  randomized experiment.
\newblock \emph{Am Econ Rev}, 77\penalty0 (3):\penalty0 251--277, Jun 1987.
\newblock ISSN 0002-8282 (Print); 0002-8282 (Linking).

\bibitem[Marx and Vreeken(2019)]{Marx2019}
Alexander Marx and Jilles Vreeken.
\newblock Causal inference on multivariate and mixed-type data.
\newblock In Michele Berlingerio, Francesco Bonchi, Thomas G{\"a}rtner, Neil
  Hurley, and Georgiana Ifrim, editors, \emph{Machine Learning and Knowledge
  Discovery in Databases}, pages 655--671, Cham, 2019. Springer International
  Publishing.
\newblock ISBN 978-3-030-10928-8.

\bibitem[Mooij et~al.(2016)Mooij, Peters, Janzing, Zscheischler, and
  Sch{{\"o}}lkopf]{JMLR:v17:14-518}
Joris~M. Mooij, Jonas Peters, Dominik Janzing, Jakob Zscheischler, and Bernhard
  Sch{{\"o}}lkopf.
\newblock Distinguishing cause from effect using observational data: Methods
  and benchmarks.
\newblock \emph{Journal of Machine Learning Research}, 17\penalty0
  (32):\penalty0 1--102, 2016.
\newblock URL \url{http://jmlr.org/papers/v17/14-518.html}.

\bibitem[Pearl(2000)]{pearl2000}
Judea Pearl.
\newblock \emph{Causality: models, reasoning and inference}.
\newblock Cambridge University Press, 2000.

\bibitem[Pencina et~al.(2009)Pencina, D'Agostino, Larson, Massaro, and
  Vasan]{Pencina:2009aa}
Michael~J Pencina, Ralph B~Sr D'Agostino, Martin~G Larson, Joseph~M Massaro,
  and Ramachandran~S Vasan.
\newblock Predicting the 30-year risk of cardiovascular disease: the framingham
  heart study.
\newblock \emph{Circulation}, 119\penalty0 (24):\penalty0 3078--3084, Jun 2009.
\newblock ISSN 1524-4539 (Electronic); 0009-7322 (Print); 0009-7322 (Linking).
\newblock \doi{10.1161/CIRCULATIONAHA.108.816694}.

\bibitem[Peters et~al.(2017)Peters, Janzing, and Sch{\"o}lkopf]{PetJanSch17}
J.~Peters, D.~Janzing, and B.~Sch{\"o}lkopf.
\newblock \emph{Elements of Causal Inference - Foundations and Learning
  Algorithms}.
\newblock Adaptive Computation and Machine Learning Series. The MIT Press,
  Cambridge, MA, USA, 2017.
\newblock URL
  \url{https://library.oapen.org/bitstream/id/056a11be-ce3a-44b9-8987-a6c68fce8d9b/11283.pdf}.

\bibitem[Sch\"{o}lkopf et~al.(2012)Sch\"{o}lkopf, Janzing, Peters, Sgouritsa,
  Zhang, and Mooij]{10.5555/3042573.3042635}
Bernhard Sch\"{o}lkopf, Dominik Janzing, Jonas Peters, Eleni Sgouritsa, Kun
  Zhang, and Joris Mooij.
\newblock On causal and anticausal learning.
\newblock In \emph{Proceedings of the 29th International Coference on
  International Conference on Machine Learning}, ICML'12, pages 459--466,
  Madison, WI, USA, 2012. Omnipress.
\newblock ISBN 9781450312851.

\bibitem[Spirtes and Glymour(1991)]{Spirtes91}
Peter Spirtes and Clark Glymour.
\newblock An algorithm for fast recovery of sparse causal graphs.
\newblock \emph{Social Science Computer Review}, 9\penalty0 (1):\penalty0
  62--72, 1991.

\bibitem[Spirtes et~al.(1999)Spirtes, Meek, and Richardson]{fci}
Peter Spirtes, Christopher Meek, and Thomas Richardson.
\newblock Causal discovery in the presence of latent variables and selection
  bias.
\newblock In Gregory~Floyd Cooper and Clark~N Glymour, editors,
  \emph{Computation, Causality, and Discovery}, pages 211--252. AAAI Press,
  1999.

\bibitem[Spirtes et~al.(2000)Spirtes, Glymour, Scheines, Heckerman, Meek,
  Cooper, and Richardson]{spirtes2000}
Peter Spirtes, Clark~N Glymour, Richard Scheines, David Heckerman, Christopher
  Meek, Gregory Cooper, and Thomas Richardson.
\newblock \emph{Causation, prediction, and search}.
\newblock MIT press, 2000.

\bibitem[Tsagris et~al.(2018)Tsagris, Borboudakis, Lagani, and
  Tsamardinos]{Tsagris:2018aa}
Michail Tsagris, Giorgos Borboudakis, Vincenzo Lagani, and Ioannis Tsamardinos.
\newblock Constraint-based causal discovery with mixed data.
\newblock \emph{International Journal of Data Science and Analytics},
  6\penalty0 (1):\penalty0 19--30, 2018.
\newblock \doi{10.1007/s41060-018-0097-y}.
\newblock URL \url{https://doi.org/10.1007/s41060-018-0097-y}.

\bibitem[Wei et~al.(2011)Wei, Coady, Goff, Brancati, Levy, Selvin, Vasan, and
  Fox]{Wei:2011aa}
Gina~S Wei, Sean~A Coady, David C~Jr Goff, Frederick~L Brancati, Daniel Levy,
  Elizabeth Selvin, Ramachandran~S Vasan, and Caroline~S Fox.
\newblock Blood pressure and the risk of developing diabetes in african
  americans and whites: Aric, cardia, and the framingham heart study.
\newblock \emph{Diabetes Care}, 34\penalty0 (4):\penalty0 873--879, Apr 2011.
\newblock ISSN 1935-5548 (Electronic); 0149-5992 (Print); 0149-5992 (Linking).
\newblock \doi{10.2337/dc10-1786}.

\bibitem[Wenjuan et~al.(2018)Wenjuan, Lu, and Chunchen]{ijcai2018p711}
Wei Wenjuan, Feng Lu, and Liu Chunchen.
\newblock Mixed causal structure discovery with application to prescriptive
  pricing.
\newblock In \emph{Proceedings of the Twenty-Seventh International Joint
  Conference on Artificial Intelligence, {IJCAI-18}}, pages 5126--5134.
  International Joint Conferences on Artificial Intelligence Organization, 7
  2018.
\newblock \doi{10.24963/ijcai.2018/711}.
\newblock URL \url{https://doi.org/10.24963/ijcai.2018/711}.

\bibitem[Willis and Rosen(1979)]{RePEc:ucp:jpolec:v:87:y:1979:i:5:p:s7-36}
Robert Willis and Sherwin Rosen.
\newblock Education and self-selection.
\newblock \emph{Journal of Political Economy}, 87\penalty0 (5):\penalty0
  S7--36, 1979.
\newblock URL
  \url{https://EconPapers.repec.org/RePEc:ucp:jpolec:v:87:y:1979:i:5:p:s7-36}.

\bibitem[Yao et~al.(2025)Yao, Verdonck, and Raymaekers]{pmlr-v258-yao25a}
Ruicong Yao, Tim Verdonck, and Jakob Raymaekers.
\newblock Causal discovery in mixed additive noise models.
\newblock In Yingzhen Li, Stephan Mandt, Shipra Agrawal, and Emtiyaz Khan,
  editors, \emph{Proceedings of The 28th International Conference on Artificial
  Intelligence and Statistics}, volume 258 of \emph{Proceedings of Machine
  Learning Research}, pages 3088--3096. PMLR, 03--05 May 2025.
\newblock URL \url{https://proceedings.mlr.press/v258/yao25a.html}.

\bibitem[Zeng et~al.(2022)Zeng, Shimizu, Matsui, and Sun]{pmlr-v177-zeng22a}
Yan Zeng, Shohei Shimizu, Hidetoshi Matsui, and Fuchun Sun.
\newblock Causal discovery for linear mixed data.
\newblock In Bernhard Sch{\"o}lkopf, Caroline Uhler, and Kun Zhang, editors,
  \emph{Proceedings of the First Conference on Causal Learning and Reasoning},
  volume 177 of \emph{Proceedings of Machine Learning Research}, pages
  994--1009. PMLR, 11--13 Apr 2022.
\newblock URL \url{https://proceedings.mlr.press/v177/zeng22a.html}.

\end{thebibliography}

\newpage

\onecolumn

\title{Density Ratio-based Causal Discovery\\ from Bivariate Continuous-Discrete Data\\(Supplementary Material)}
\maketitle

\begin{center}\vspace{3em}
\Large\textbf{Contents}
\end{center}

\vspace{1em}

\begingroup
\let\clearpage\relax
\let\cleardoublepage\relax

\textbf{A\ \ Proofs} \dotfill \pageref{sec:proofs}
\begin{itemize}
  \item A.1\quad Proof of Lemma~1 \dotfill \pageref{sec:lem1}
  \item A.2\quad Proof of Lemma~2 \dotfill \pageref{sec:lem2}
  \item A.3\quad Proof of Corollary~1 \dotfill \pageref{sec:cor1}
  \item A.4\quad Proof of Lemma~3 \dotfill \pageref{sec:lem3}
  \item A.5\quad Proof of Lemma~4 \dotfill \pageref{sec:lem4}
  \item A.6\quad Proof of Theorem~1 \dotfill \pageref{sec:thm1}
\end{itemize}

\vspace{0.5em}

\textbf{B\ \ Computational Complexity} \dotfill \pageref{sec:complexity}

\vspace{0.5em}

\textbf{C\ \ Illustrative example settings} \dotfill \pageref{sec:example1}

\vspace{0.5em}

\textbf{D\ \ Details of the experiments} \dotfill \pageref{sec:expedetails}
\begin{itemize}
  \item D.1\quad Code availability \dotfill \pageref{sec:codeavail}
  \item D.2\quad Code packages and implementations \dotfill \pageref{sec:packages}
  \item D.3\quad Experimental settings \dotfill \pageref{sec:settingsex}
  \item D.4\quad Computational resources \dotfill \pageref{sec:resources}
\end{itemize}

\vspace{0.5em}

\textbf{E\ \ Causal Evidence from Epidemiological Studies} \dotfill \pageref{sec:causal_evidence}

\vspace{0.5em}

\textbf{F\ \ Sensitivity Analysis} \dotfill \pageref{sec:sensitivity}
\begin{itemize}
  \item F.1\quad Sensitivity to hyperparameters $\rho$ and $\gamma$ \dotfill \pageref{sec:sensitivity_hyperparam}
  \item F.2\quad Sensitivity to number of observations \dotfill \pageref{sec:sensitivity_samplesize}
\end{itemize}

\endgroup

\clearpage

\appendix

\section{Proofs}
\label{sec:proofs}

\subsection{Proof of Lemma~\ref{lem:location-shift-rare}}
\label{sec:lem1}

\textbf{Lemma~\ref{lem:location-shift-rare}.} \textit{
Assume that $(X,Y)$ are generated according to the model in Section~\ref{sec:causal-model} with causal direction $X \to Y$. Let $c_s$ and $c_t$ be distinct outcomes of $Y$, and let $i$ be an index such that $Z_i$ separates $c_s$ and $c_t$ as guaranteed by Assumption~\ref{assumption:separability}. Let $F_i(\cdot)$ denote the cumulative distribution function (CDF) of the noise variable $N_{Y,i}$. If the conditional distributions $P(X|Y=c_s)$ and $P(X|Y=c_t)$ belong to the same location-shift family with shift $\mu \neq 0$, then the mechanism function $f_i$, the input density $p_X$, and the noise CDF $F_i$ must satisfy
\begin{align*}
\frac{1 - F_i(-f_i(x))}{F_i(-f_i(x - \mu))} = C \cdot \frac{p_X(x - \mu)}{p_X(x)}
\end{align*}
for all $x$, where $C > 0$ is a constant determined by the marginal probabilities of $Y$.
}

\begin{proof}
Recall that in our model (Section~\ref{sec:causal-model}), the output $Y$ is generated through the function $Y = q((Z_i)_{i=1}^m)$, where each $Z_i = \mathbb{I}[f_i(X) + N_{Y,i} \geq 0]$. 

By Assumption~\ref{assumption:separability}, there exists an indicator $Z_i = \mathbb{I}[f_i(X) + N_{Y,i} \ge 0]$ that distinguishes $c_s$ and $c_t$. Without loss of generality, assume $Z_i=0$ for $Y=c_s$ and $Z_i=1$ for $Y=c_t$.

This means that the following equivalences hold:
\begin{align}
P(Y=c_s\mid X=x,Y\in \{c_s,c_t\})=P(f_i(x) + N_{Y,i} < 0 \mid X=x),
\end{align}
and
\begin{align}
\begin{aligned}
P(Y=c_t\mid X=x,Y\in \{c_s,c_t\})
&=1-P(Y=c_s\mid X=x,Y\in \{c_s,c_t\})\\
&=1-P(f_i(x) + N_{Y,i} < 0 \mid X=x).
\end{aligned}
\end{align}

Using Bayes' rule, the conditional density for $X$ given $Y=c_s$ can be written as
\begin{align}
\begin{aligned}
p_{X\mid Y=c_s}(x)
&=\frac{P(Y=c_s\mid X=x)p_X(x)}{p_Y(c_s)} \\
&=\frac{P(Y=c_s\mid X=x,Y\in \{c_s,c_t\})p_X(x)P(Y\in\{c_s,c_t\})}{p_Y(c_s)}\\
&=\frac{P(Y=c_s\mid X=x,Y\in \{c_s,c_t\})p_X(x)\left(p_Y(c_s)+p_Y(c_t)\right)}{p_Y(c_s)}\\
&=\frac{P(f_i(x) + N_{Y,i} < 0 \mid X=x)p_X(x)\left(p_Y(c_s)+p_Y(c_t)\right)}
{p_Y(c_s)}.
\end{aligned}
\end{align}

The conditional density for $X$ given $Y=c_t$ can be written as
\begin{align}
\begin{aligned}
p_{X\mid Y=c_t}(x)
&=\frac{P(Y=c_t\mid X=x)p_X(x)}{p_Y(c_t)} \\
&=\frac{P(Y=c_t\mid X=x,Y\in \{c_s,c_t\})p_X(x)\left(p_Y(c_s)+p_Y(c_t)\right)}{p_Y(c_t)}\\
&=\frac{\Big(1-P(f_i(x) + N_{Y,i} < 0\mid X=x)\Big)p_X(x)\left(p_Y(c_s)+p_Y(c_t)\right)}{p_Y(c_t)}.
\end{aligned}
\end{align}

Since $N_{Y,i}$ is independent of $X$, we have 
\begin{align}
\begin{aligned}
P(f_i(x) + N_{Y,i} < 0 \mid X=x)
&=\int_{-\infty}^{\,-f_i(x)} p_{N_{Y,i}}(u)\,du\\
&= F_i(-f_i(x)).
\end{aligned}
\end{align}

Then, the conditional densities become
\begin{align}
p_{X\mid Y=c_s}(x)
&=\frac{F_i(-f_i(x))}{p_Y(c_s)}p_X(x)(p_Y(c_s)+p_Y(c_t)),\label{eq:cond_density_cs}
\end{align}
and
\begin{align}
p_{X\mid Y=c_t}(x)=\frac{1-F_i(-f_i(x))}{p_Y(c_t)}p_X(x)(p_Y(c_s)+p_Y(c_t)).
\label{eq:cond_density_ct}
\end{align}

The assumption that the distributions form a location-shift family implies:
\begin{align}
p_{X\mid Y=c_t}(x) = p_{X\mid Y=c_s}(x-\mu).
\end{align}
Substituting equation~(\ref{eq:cond_density_ct}) into the left-hand side and equation~(\ref{eq:cond_density_cs}) with $x$ replaced by $x - \mu$ into the right-hand side, and canceling the common factor $(p_Y(c_s)+p_Y(c_t))$:
\begin{align}
\frac{(1 - F_i(-f_i(x))) p_X(x)}{p_Y(c_t)} = \frac{F_i(-f_i(x-\mu)) p_X(x-\mu)}{p_Y(c_s)}.
\end{align}
Rearranging the terms to separate the functional components:
\begin{align}
\frac{1 - F_i(-f_i(x))}{F_i(-f_i(x-\mu))} = \frac{p_Y(c_t)}{p_Y(c_s)} \cdot \frac{p_X(x-\mu)}{p_X(x)}.
\end{align}
Letting $C = \frac{p_Y(c_t)}{p_Y(c_s)}$, we obtain the condition stated in the lemma.
\end{proof}

\subsection{Proof of Lemma~\ref{lem:monotonicity-conditions}}
\label{sec:lem2}

\textbf{Lemma~\ref{lem:monotonicity-conditions}.} {\it
Let $f$ and $g$ be finite mixtures of generalized normal distributions:
\begin{align*}
f(x) &= \sum_{j=1}^{L_f} \omega^f_j \, \varphi(x;\, \mu^f_j, \alpha^f_j, \beta^f_j), \\
g(x) &= \sum_{j=1}^{L_g} \omega^g_j \, \varphi(x;\, \mu^g_j, \alpha^g_j, \beta^g_j),
\end{align*}
where $\varphi(x;\, \mu, \alpha, \beta) = \frac{\beta}{2\alpha\Gamma(1/\beta)} \exp\!\left(-\left|\frac{x-\mu}{\alpha}\right|^\beta\right)$ is the generalized normal density with location $\mu \in \mathbb{R}$, scale $\alpha > 0$, and shape $\beta > 0$. We parameterize the mixture weights by their first $L_f-1$ (resp.\ $L_g-1$) components, with $\omega^f_{L_f} = 1 - \sum_{j=1}^{L_f-1}\omega^f_j$ (resp.\ $\omega^g_{L_g} = 1 - \sum_{j=1}^{L_g-1}\omega^g_j$), subject to all weights being strictly positive. Write $\theta_f = (\omega^f_{1:L_f-1},\, \mu^f,\, \alpha^f,\, \beta^f)$ and $\theta_g = (\omega^g_{1:L_g-1},\, \mu^g,\, \alpha^g,\, \beta^g)$ for the parameter vectors of $f$ and $g$, respectively. Define the following two parameter spaces:
\begin{itemize}
\item[\textup{(a)}] $\Theta_a$: all entries of $(\theta_f, \theta_g)$ are free, so that $\Theta_a$ is an open subset of $\mathbb{R}^{4L_f+4L_g-2}$.
\item[\textup{(b)}] $\Theta_b$: the shape parameters $\beta^f$ and $\beta^g$ are fixed to arbitrary positive constants and all remaining entries of $(\theta_f, \theta_g)$ are free, so that $\Theta_b$ is an open subset of $\mathbb{R}^{3L_f+3L_g-2}$.
\end{itemize}
Let $\mathcal{M}_a = \{\theta \in \Theta_a : r(x) = f(x)/g(x) \text{ is monotonic}\}$ and $\mathcal{M}_b = \{\theta \in \Theta_b : r(x) = f(x)/g(x) \text{ is monotonic}\}$. Then $\mathcal{M}_a$ has Lebesgue measure zero in $\Theta_a$, and $\mathcal{M}_b$ has Lebesgue measure zero in $\Theta_b$.
}

\begin{proof}
The proof proceeds as follows. First, in the Preliminary, we analyze the tail asymptotics of generalized normal mixtures and define the notion of a tail-dominant component. Then, in Step~1, we show that the density ratio $r(x) = f(x)/g(x)$ is non-monotonic whenever the tail-dominant components of $f$ and $g$ differ in shape or scale. Finally, in Step~2, we use Step~1 to show that $\mathcal{M}_a$ and $\mathcal{M}_b$ each have Lebesgue measure zero.

\textbf{Preliminary: Tail asymptotics.}
The tail behavior of each generalized normal component is governed by its shape and scale parameters: as $|x| \to \infty$,
\begin{align}
\log \varphi(x;\, \mu_j, \alpha_j, \beta_j) = -\frac{|x - \mu_j|^{\beta_j}}{\alpha_j^{\beta_j}} + \mathrm{const} \sim -\frac{|x|^{\beta_j}}{\alpha_j^{\beta_j}}.
\end{align}
Since $|x|^{\beta}$ is symmetric, the same component dominates in both tails ($x \to +\infty$ and $x \to -\infty$). Among all components in a mixture, the one with the smallest $\beta_j$ has the heaviest tail; if several components share the same smallest $\beta_j$, the one with the largest $\alpha_j$ decays most slowly. Note that the mixture weights do not affect the tail asymptotics as long as they are strictly positive, which is guaranteed by our parameterization.

We call such a component a \emph{tail-dominant component} of the mixture. Formally, for the mixture $f = \sum_{j} \omega^f_j \varphi(x;\, \mu^f_j, \alpha^f_j, \beta^f_j)$, a tail-dominant component is a component $j^*$ that achieves $\beta^f_{j^*} = \min_j \beta^f_j$ and, among all indices attaining this minimum, $\alpha^f_{j^*} = \max \{ \alpha^f_j : \beta^f_j = \beta^f_{j^*} \}$. We define $\beta^f_{\min} = \beta^f_{j^*}$ and $\alpha_f = \alpha^f_{j^*}$, and analogously $\beta^g_{\min}$, $\alpha_g$ for the mixture $g$.

\textbf{Step 1: Non-monotonicity from tail mismatch.}
We show that the density ratio $r(x) = f(x)/g(x)$ is non-monotonic whenever $\beta^f_{\min} \neq \beta^g_{\min}$ or $\alpha_f \neq \alpha_g$, i.e., whenever the tail-dominant components of $f$ and $g$ differ in shape or scale. To see this, we consider two cases.

\textbf{Step 1a: $\beta^f_{\min} \neq \beta^g_{\min}$.} Without loss of generality, suppose $\beta^f_{\min} < \beta^g_{\min}$. Then $f$ has heavier tails than $g$, and as $|x| \to \infty$,
\begin{align}
\log r(x) = \log f(x) - \log g(x) \to +\infty.
\end{align}
Since this holds in both tails, $r(x) \to +\infty$ as $x \to +\infty$ and as $x \to -\infty$. Because $r$ is continuous and positive, it attains a minimum at some interior point $x_1$. Then there exist $x_0 < x_1 < x_2$ such that $r(x_0) > r(x_1)$ and $r(x_1) < r(x_2)$, i.e., $(r(x_0) - r(x_1))(r(x_1) - r(x_2)) < 0$, so $r$ is non-monotonic (Definition~\ref{defn:non-monotonic}).

\textbf{Step 1b: $\beta^f_{\min} = \beta^g_{\min}$ and $\alpha_f \neq \alpha_g$.} As $|x| \to \infty$,
\begin{align}
\log r(x) \sim \left(\frac{1}{\alpha_g^{\beta_{\min}}} - \frac{1}{\alpha_f^{\beta_{\min}}}\right)|x|^{\beta_{\min}},
\end{align}
which diverges to $+\infty$ in both tails or to $-\infty$ in both tails, depending on the sign of the coefficient. In either case, because $r$ is continuous and positive, it attains an extremum at some interior point $x_1$. Then there exist $x_0 < x_1 < x_2$ such that $(r(x_0) - r(x_1))(r(x_1) - r(x_2)) < 0$, so $r$ is non-monotonic (Definition~\ref{defn:non-monotonic}).

\textbf{Step 2: $\mathcal{M}_a$ and $\mathcal{M}_b$ have Lebesgue measure zero.}
We treat the two parameter spaces separately.

\textbf{Case~(a):} In $\Theta_a$, all parameters including the shape parameters $\beta^f_j$ and $\beta^g_k$ are free. By Step~1a, $r(x)$ is non-monotonic whenever $\beta^f_{\min} \neq \beta^g_{\min}$, so $\mathcal{M}_a \subseteq \{\theta \in \Theta_a : \beta^f_{\min} = \beta^g_{\min}\}$. If $\beta^f_{\min} = \beta^g_{\min}$, then the minimum of $\{\beta^f_j\}$ equals the minimum of $\{\beta^g_k\}$, so there must exist indices $j_0 \in \{1,\ldots,L_f\}$ and $k_0 \in \{1,\ldots,L_g\}$ such that $\beta^f_{j_0} = \beta^g_{k_0}$. Therefore $\{\beta^f_{\min} = \beta^g_{\min}\} \subseteq \bigcup_{j_0,k_0}\{\beta^f_{j_0} = \beta^g_{k_0}\}$, which is a finite union of codimension-one sets in $\Theta_a$, and hence $\mathcal{M}_a$ has Lebesgue measure zero in $\Theta_a$.

\textbf{Case~(b):} In $\Theta_b$, the shape parameters are fixed constants and the remaining parameters are free. If the fixed values satisfy $\beta^f_{\min} \neq \beta^g_{\min}$, then by Step~1a, $r(x)$ is non-monotonic for all $\theta \in \Theta_b$, so $\mathcal{M}_b = \emptyset$. If $\beta^f_{\min} = \beta^g_{\min}$, then by Step~1b, monotonicity further requires $\alpha_f = \alpha_g$, so $\mathcal{M}_b \subseteq \{\theta \in \Theta_b : \alpha_f = \alpha_g\}$. Since the scale parameters are free in $\Theta_b$, $\{\alpha_f = \alpha_g\} \subseteq \bigcup_{j_0, k_0} \{\alpha^f_{j_0} = \alpha^g_{k_0}\}$ for $j_0 \in \{j : \beta^f_j = \beta^f_{\min}\}$ and $k_0 \in \{k : \beta^g_k = \beta^g_{\min}\}$, which is a finite union of codimension-one sets in $\Theta_b$. Therefore $\mathcal{M}_b$ has Lebesgue measure zero in $\Theta_b$.

In both cases, $\mathcal{M}_a$ and $\mathcal{M}_b$ have Lebesgue measure zero in their respective parameter spaces, completing the proof.
\end{proof}

\noindent\textbf{Remark.}
The non-genericity of monotonic density ratios likely extends beyond the generalized normal mixture family: when two conditional distributions arise from independent mechanisms, their density ratio satisfying monotonicity is highly unlikely. Formal measure-zero results, however, require a concrete parameterization over which the measure is defined, for which generalized normal mixtures provide a flexible yet tractable choice.

\subsection{Proof of Corollary~\ref{cor:nonmonotonic}}
\label{sec:cor1}

{\bf Corollary~\ref{cor:nonmonotonic}:} {\it
Under the model $Y \to X$ with non-location-shift conditionals (Case~2 in Section~\ref{sec:model-YX}), the density ratio $G_{c_s,c_t}(x)$ is monotonic only on a set of Lebesgue measure zero in the parameter space.
}

\begin{proof}
Under the model $Y \to X$ with non-location-shift conditionals (Case~2 in Section~\ref{sec:model-YX}), the conditional distributions $P(X|Y=c_s)$ and $P(X|Y=c_t)$ are finite mixtures of generalized normal distributions whose parameters are independently parameterized. This corresponds directly to case~(a) of Lemma~\ref{lem:monotonicity-conditions}, with $f = p_{X|Y=c_t}$ and $g = p_{X|Y=c_s}$. By Lemma~\ref{lem:monotonicity-conditions}, the set of parameters for which the density ratio $G_{c_s,c_t}(x) = f(x)/g(x)$ is monotonic has Lebesgue measure zero in $\Theta_a$.
\end{proof}

\subsection{Proof of Lemma~\ref{lem:main}}
\label{sec:lem3}

{\bf Lemma~\ref{lem:main}:} {\it
Assume that $(X,Y)$ are generated according to the model described in Section~\ref{sec:causal-model} with the causal direction $X\to Y$. Then, for any distinct outcomes $c_s$ and $c_t$ of $Y$, the density ratio $G_{c_s,c_t}(x)$ is monotonic in $x$.
}

\begin{proof}
By Assumption~\ref{assumption:separability}, there exists an indicator $Z_i = \mathbb{I}[f_i(X) + N_{Y,i} \ge 0]$ that distinguishes $c_s$ and $c_t$. By the same derivation as in the proof of Lemma~\ref{lem:location-shift-rare} (see Equations~\eqref{eq:cond_density_cs} and~\eqref{eq:cond_density_ct}), the conditional densities are given by
\begin{align}
p_{X\mid Y=c_s}(x)
&=\frac{F_i(-f_i(x))}{p_Y(c_s)}p_X(x)(p_Y(c_s)+p_Y(c_t)),\label{eq:cond_density_aacs}
\end{align}
and
\begin{align}
p_{X\mid Y=c_t}(x)=\frac{1-F_i(-f_i(x))}{p_Y(c_t)}p_X(x)(p_Y(c_s)+p_Y(c_t)),
\label{eq:cond_density_aact}
\end{align}
where $F_i(\cdot)$ denotes the cumulative distribution function of the noise $N_{Y,i}$.

We let $C$ denote a constant as $C=\frac{p_Y(c_s)}{p_Y(c_t)}$. Combining \eqref{eq:cond_density_aacs} and \eqref{eq:cond_density_aact}, the density ratio becomes
\begin{align}\label{eq:ratio_final}
G_{c_s,c_t}(x)
&=\frac{p_{X\mid Y=c_t}(x)}{p_{X\mid Y=c_s}(x)}
=C\cdot\frac{1-F_i(-f_i(x))}{F_i(-f_i(x))}=C \cdot \left( \frac{1}{F_i(-f_i(x))} - 1 \right).
\end{align}
Since $f_i(x)$ is monotonic and $F_i$ is a CDF (monotonically increasing), the composite function $F_i(-f_i(x))$ is monotonic. Consequently, $G_{c_s,c_t}(x)$ is monotonic in $x$.
\end{proof}

\subsection{Proof of Lemma~\ref{lem:no_causal}}
\label{sec:lem4}

\textbf{Lemma~\ref{lem:no_causal}.} \textit{
The conditional distributions $P_{X|Y}(x|c_s)$ and $P_{X|Y}(x|c_t)$ are identical for all distinct $c_s$ and $c_t$ if and only if $X$ and $Y$ have no causal relationship. Equivalently, $G_{c_s,c_t}(x) = 1$ for all $x$ if and only if $X$ and $Y$ are independent.
}

\begin{proof}
By the definition of independence, $X$ and $Y$ are independent if and only if the joint density factorizes as $p_{X,Y}(x, c) = p_X(x) p_Y(c)$ for all $x$ and $c$. This holds if and only if the conditional density satisfies $p_{X \mid Y}(x \mid c) = p_X(x)$ for all $x$ and all values $c$ of $Y$. In particular, this implies $p_{X \mid Y}(x \mid c_s) = p_{X \mid Y}(x \mid c_t) = p_X(x)$ for any distinct $c_s$ and $c_t$, and therefore $G_{c_s,c_t}(x) = 1$ for all $x$.

Conversely, if $G_{c_s,c_t}(x) = 1$ for all pairs of distinct values $c_s$ and $c_t$, then $p_{X \mid Y}(x \mid c)$ is the same for all $c$. Since $p_X(x) = \sum_c p_{X \mid Y}(x \mid c) p_Y(c)$, this common conditional density must equal $p_X(x)$, which implies independence.
\end{proof}

\subsection{Proof of Theorem~\ref{thm:identifiability}}
\label{sec:thm1}

{\bf Theorem~\ref{thm:identifiability}:} {\it
Assume that the joint distribution of $(X,Y)$ is generated according to exactly one of the three causal models defined in Section~\ref{sec:model}: (1) $X \to Y$, (2) $Y \to X$, or (3) no causal relationship. Then the causal relationship between $X$ and $Y$ is identifiable.}

\begin{proof}
We show that the three causal models are mutually distinguishable.

By Lemma~\ref{lem:no_causal}, when there is no causal relationship, $G_{c_s,c_t}(x) = 1$ for all $x$, which distinguishes this case from the other two.

By Lemma~\ref{lem:main}, the density ratio $G_{c_s,c_t}(x)$ is monotonic under $X \to Y$, and by Assumption~\ref{assumption:non-location-shift}, the conditional distributions do not form a location-shift family.

Under $Y \to X$, the conditionals either form a location-shift family or do not. In the former case, the direction is identified because location-shift conditionals cannot arise under $X \to Y$ (Assumption~\ref{assumption:non-location-shift}). In the latter case, by Corollary~\ref{cor:nonmonotonic}, the density ratio is non-monotonic except on a set of Lebesgue measure zero in the parameter space, which is incompatible with the monotonicity established under $X \to Y$.

Thus, all three causal models are distinguishable, and the causal relationship is identifiable.
\end{proof}

\section{Computational Complexity}
\label{sec:complexity}
We analyze the time and memory complexity of DRCD by considering each step of Algorithm~\ref{algo:drcd}. 
Throughout this section, let $n$ be the total number of observations, $c_s$ and $c_t$ be the two most frequent values of $Y$, 
$n_{s}$ and $n_{t}$ be the numbers of samples with $Y=c_s$ and $Y=c_t$, respectively, 
$b$ the number of kernel basis functions (centers) used in uLSIF, 
$S$ and $L$ the numbers of candidate kernel widths and regularization parameters (hyperparameters for uLSIF), 
and $n_{\text{points}}$ the number of grid points used in monotonicity testing.

\paragraph{Step 1: Testing for causal existence.}
We perform a two-sample KS test between the conditional samples $\{x \mid Y=c_s\}$ and $\{x \mid Y=c_t\}$. Computing empirical distribution functions requires sorting and thus $O(n\log n)$ time. The memory usage of this step is $O(n)$.

\paragraph{Step 2: Testing for location-shift.}
We perform a centered KS test by first centering each conditional sample (subtracting its mean) and then applying the two-sample KS test. Computing the means requires $O(n)$ time, and the subsequent KS test requires $O(n\log n)$ time for sorting. The memory usage of this step is $O(n)$.

\paragraph{Step 3: Density-ratio estimation (uLSIF).}
For each pair $(c_s,c_t)$, we estimate the density ratio $G_{c_s,c_t}(x)$ using uLSIF with a fixed kernel width $\sigma$ and regularization parameter $\lambda$. The parameters are obtained by minimizing
\begin{equation}
J(\beta)=\frac{1}{2}\beta^{\top}\hat{H}\beta - \hat{h}^{\top}\beta, \qquad \hat{\beta}=(\hat{H}+\lambda I)^{-1}\hat{h},
\end{equation}
where $\hat{H}\in\mathbb{R}^{b\times b}$ and $\hat{h}\in\mathbb{R}^{b}$. The computational cost of one solution is: (i) computing $\hat{H}$: $O(n_{s}b^{2})$ (outer products of basis functions); (ii) computing $\hat{h}$: $O(n_{t}b)$; (iii) solving the linear system: $O(b^{3})$ (e.g., Cholesky decomposition). Because the leave-one-out cross-validation (LOOCV) score can be computed analytically, the cost of model selection over all $S$ kernel widths and $L$ regularization parameters remains
\begin{equation}
O\bigl(SL(n_{s}b^{2}+n_{t}b+b^{3})\bigr).
\end{equation}
Memory usage is dominated by storing $\hat{H}$, requiring $O(b^{2})$ space.

\paragraph{Step 4: Monotonicity evaluation.}
Let $x_{\min}$ and $x_{\max}$ denote the lower and upper bounds of the overlapping support of $\{x \mid Y=c_s\}$ and $\{x \mid Y=c_t\}$. Evaluating the density-ratio estimate $G_{c_s,c_t}(x)$ at $n_{\text{points}}$ grid points requires $O(n_{\text{points}}b)$ operations (kernel expansions). We then compute the Spearman rank correlation coefficient between the indices $(1, 2, \ldots, n_{\text{points}})$ and the density ratio values. Since the indices are already sorted, computing the Spearman correlation requires only ranking the density ratio values, which takes $O(n_{\text{points}}\log n_{\text{points}})$ time for sorting, followed by $O(n_{\text{points}})$ time to compute the correlation. The memory usage of this step is $O(n_{\text{points}})$, for storing the grid values and their ranks.

\paragraph{Overall complexity.}
Combining the four steps, the overall time complexity is
\begin{equation}
O\bigl(n\log n + SL(n_{s}b^{2}+n_{t}b+b^{3}) + n_{\text{points}}b + n_{\text{points}}\log n_{\text{points}}\bigr),
\end{equation}
where all symbols are as defined above. In practice, $n_{\text{points}}\log n_{\text{points}}$ is dominated by the other terms. The memory complexity is
\begin{equation}
O(n+b^{2}+n_{\text{points}}).
\end{equation}
In practice, the number of basis functions is typically on the order of $b \approx 100$, 
so the $O(b^{2})$ term dominates the algorithm-specific memory usage.

\section{Illustrative example settings}
\label{sec:example1}

This appendix explains the generative models used to produce the illustrative plots in Figure~\ref{figure:plot}.
Note that Gaussianity is assumed in these examples solely for illustrative purposes; our theoretical framework does not require such distributional assumptions.

\subsubsection*{Case 1: $X \rightarrow Y$}

In this case, the continuous variable $X$ causes the binary variable $Y$ through a noisy threshold model. The structural equations are modeled as:
\begin{align*}
    X &= \xi_X, \quad \xi_X \sim \mathcal{N}(0, 1), \\
    Y &= 
    \begin{cases}
    1 & \text{if } X + \xi_Y > 0, \\
    0 & \text{otherwise},
    \end{cases}
    \quad \xi_Y \sim \mathcal{N}(0, 1).
\end{align*}

Under this model, the distribution of $X$ is:
\begin{align*}
P(X) = \mathcal{N}(X; 0, 1).
\end{align*}

The conditional densities of $X$ given $Y$ are:
\begin{align*}
P(X \mid Y=1) &= \frac{\Phi(X) \cdot \mathcal{N}(X; 0, 1)}{p_Y(1)}, \\
P(X \mid Y=0) &= \frac{(1 - \Phi(X)) \cdot \mathcal{N}(X; 0, 1)}{p_Y(0)},
\end{align*}
where $\Phi(\cdot)$ denotes the standard Gaussian cumulative distribution function, and $p_Y(1)=p_Y(0)=0.5$ holds due to the symmetry of this model. 

The resulting density ratio is:
\begin{align*}
\frac{P(X \mid Y=1)}{P(X \mid Y=0)} = \frac{\Phi(X)}{1 - \Phi(X)}.
\end{align*}

\subsubsection*{Case 2: $Y \rightarrow X$}

In this case, the binary variable $Y$ causes the continuous variable $X$ through a Gaussian mixture model. The structural equations are modeled as:
\begin{align*}
    Y &\sim \text{Bernoulli}(0.5), \\
    X &= 
    \begin{cases}
    \xi_1, & \text{if } Y = 1, \\
    \xi_2, & \text{if } Y = 0,
    \end{cases}
    \quad \xi_1 \sim \mathcal{N}(0.5, 0.25^2), \quad \xi_2 \sim \mathcal{N}(-0.5, 0.5^2).
\end{align*}

The conditional distributions of $X$ given $Y$ are:
\begin{align*}
P(X \mid Y=1) = \mathcal{N}(X; 0.5, 0.25^2), \quad
P(X \mid Y=0) = \mathcal{N}(X; -0.5, 0.5^2).
\end{align*}

The distribution of $X$ is:
\begin{align*}
P(X) = 0.5 \cdot \mathcal{N}(X; 0.5, 0.25^2) + 0.5 \cdot \mathcal{N}(X; -0.5, 0.5^2).
\end{align*}

The density ratio is:
\begin{align*}
\frac{P(X \mid Y=1)}{P(X \mid Y=0)} = \frac{\mathcal{N}(X; 0.5, 0.25^2)}{\mathcal{N}(X; -0.5, 0.5^2)}.
\end{align*}

\section{Details of the experiments}
\label{sec:expedetails}

\subsection{Code availability}
\label{sec:codeavail}
The code for our proposed methodology and for conducting experiments with both synthetic and real-world data is available at \url{https://github.com/causal111/DRCD}.

\subsection{Code packages and implementations}
\label{sec:packages}

For baseline methods in our experiments, we used the following existing code packages and our own implementations:

\begin{itemize}
    \item \textbf{LiM} (Causal discovery for linear mixed data)~\citep{pmlr-v177-zeng22a}
    \begin{itemize}
        \item Repository: \url{https://github.com/cdt15/lingam}
        \item Implementation language: Python
        \item License: MIT
    \end{itemize}
    
    \item \textbf{CRACK} (Classification and regression based packing of data)~\citep{Marx2019}
    \begin{itemize}
        \item Repository: \url{https://eda.rg.cispa.io/prj/crack/}
        \item Implementation language: C++
        \item License: No explicit license file found in the repository.
    \end{itemize}
    
    \item \textbf{MANMs} (Mixed additive noise models)~\citep{pmlr-v258-yao25a}
    \begin{itemize}
        \item Implementation: We developed our own Python implementation specialized for bivariate data consisting of one continuous and one discrete variable, based on the methodology described in the original paper.
    \end{itemize}
    
    \item \textbf{GSF} (Generalized Score Functions for Causal Discovery)~\citep{GSF}
    \begin{itemize}
        \item Repository: \\{\fontsize{8.1pt}{10pt}\selectfont\url{https://github.com/Biwei-Huang/Generalized-Score-Functions-for-Causal-Discovery}}
        \item Implementation language: MATLAB
        \item License: No explicit license file found in the repository.
    \end{itemize}
    \item \textbf{MIC} (Mixed causal structure discovery)~\citep{ijcai2018p711}
    \begin{itemize}
        \item Implementation: We developed our own Python implementation specialized for bivariate data consisting of one continuous and one discrete variable, based on the methodology described in the original paper.
    \end{itemize}
\end{itemize}

\subsection{Experimental settings}
\label{sec:settingsex}
For the parameter setting of the DRCD algorithm (refer to Algorithm~\ref{algo:drcd}), we set $\alpha=0.05$, $\rho=0.8$, $\gamma=0.5$, and $n_{\rm points}=1000$. The monotonicity threshold $\rho=0.8$ requires a strong monotonic association in the Spearman correlation to conclude $X \to Y$. The location-shift threshold $\gamma=0.5$ is set well above conventional significance levels because Step~2 concludes in favor of the null hypothesis rather than against it. For MIC and MANMs, we developed our own Python implementations specialized for bivariate data consisting of one continuous and one discrete variable, based on the methodologies described in the original papers. For MIC, the penalty term was set to $\lambda=0.1$. For MANMs, the significance level for independence tests was set to $0.05$. In the original experiments of MANMs, linear regression was used for linear data, while random forest regression was employed for nonlinear data. Since our synthetic experiments are based on linear data, we implemented MANMs with linear regression. For the real-world datasets, we applied the same setting. For other methods, we used the default parameter settings as provided in the code packages described in Appendix~\ref{sec:packages}.

\subsection{Computational resources}
\label{sec:resources}
All experiments were conducted on a MacBook Pro with the following specifications:
\begin{itemize}
    \item CPU: Apple M3 Max chip (16 cores)
    \item Memory: 128 GB RAM
    \item Operating System: macOS Sonoma 14.4.1
    \item Programming Environment: Python 3.11.7
\end{itemize}

\section{Causal Evidence from Epidemiological Studies}
\label{sec:causal_evidence}

We provide details on the causal relationships used as ground truth in our real-world data experiments (Section~\ref{sec:realexp}). The causal directions were determined based on two criteria: (1) inherent attributes that can only be causes, and (2) documented relationships in epidemiological studies from the Framingham Heart Study.

\paragraph{Inherent attributes.}
Two variables in Table~\ref{tab:heart_results}, \texttt{sex} and \texttt{age}, are inherent attributes that can only serve as causes, not effects of physiological measurements.

\paragraph{Sex as a cause of blood pressure and cholesterol.}
Both \citet{Pencina:2009aa} and \citet{Wei:2011aa} report systematic sex differences in blood pressure: men had higher mean systolic blood pressure than women (126 vs.\ 118 mmHg in \citet{Pencina:2009aa}; 122.7 vs.\ 114.0 mmHg in \citet{Wei:2011aa}). \citet{Pencina:2009aa} also report that men had higher total cholesterol (202 vs.\ 192 mg/dL). These findings support \texttt{sex} $\to$ \texttt{trestbps} and \texttt{sex} $\to$ \texttt{chol}.

\paragraph{Age as a cause of heart disease and diabetes.}
Both studies demonstrate that age is a predictor of disease outcomes. \citet{Pencina:2009aa} identified age as a strong predictor of 30-year CVD risk with a hazard ratio of 2.09 (95\% CI: 1.88--2.31), supporting \texttt{age} $\to$ \texttt{num}. \citet{Wei:2011aa} showed age as a predictor of diabetes incidence with hazard ratios of 1.10 (95\% CI: 0.98--1.22) and 1.08 (95\% CI: 0.99--1.18) per 5-year increase among African Americans and whites, respectively, supporting \texttt{age} $\to$ \texttt{fbs}.

\section{Sensitivity Analysis}
\label{sec:sensitivity}

We evaluate the robustness of DRCD with respect to its two key hyperparameters: the monotonicity threshold~$\rho$ and the location-shift threshold~$\gamma$ (Algorithm~\ref{algo:drcd}). We also examine how performance varies with the number of observations~$n$. All experiments use the same synthetic data generation protocol as in Section~\ref{sec:artificialexperiments}.

\subsection{Sensitivity to hyperparameters $\rho$ and $\gamma$}
\label{sec:sensitivity_hyperparam}

Table~\ref{tab:sensitivity_hyperparam} reports the accuracy of DRCD across a grid of $\rho \in \{0.6, 0.7, 0.8, 0.9\}$ and $\gamma \in \{0.3, 0.5, 0.7\}$, with $n=1000$ and $1000$ datasets per causal type.

The results reveal two interpretable trade-offs. First, the threshold~$\rho$ controls the trade-off between detecting $X \to Y$ and $Y \to X$ with non-location-shift conditionals: increasing~$\rho$ imposes a stricter monotonicity criterion, which reduces $X \to Y$ accuracy (from $95.5\%$ at $\rho=0.6$ to $83.0\%$ at $\rho=0.9$ when $\gamma=0.5$) while improving non-location-shift detection (from $70.9\%$ to $91.7\%$). Second, $\gamma$ governs the trade-off between $X \to Y$ and $Y \to X$ with location-shift conditionals: a lower~$\gamma$ makes the location-shift test more permissive, improving location-shift detection at the cost of $X \to Y$ accuracy. The accuracy for the no-causation scenario remains constant at $94.7\%$ across all settings, as expected since Step~1 is independent of both~$\rho$ and~$\gamma$.

All four causal scenarios simultaneously exceed $80\%$ accuracy when $\rho \geq 0.8$, with the exception of $(\rho, \gamma) = (0.8, 0.7)$, where the location-shift detection drops to $71.5\%$ due to the overly strict location-shift threshold. For $\rho \leq 0.7$, the non-location-shift accuracy remains below $80\%$ regardless of~$\gamma$.

\begin{table*}[h]
\centering
\caption{Sensitivity of DRCD accuracy (\%) to hyperparameters $\rho$ (monotonicity threshold) and $\gamma$ (location-shift threshold). Default setting ($\rho{=}0.8$, $\gamma{=}0.5$) is highlighted with $\dagger$. Accuracies above 80\% are shown in \textbf{bold}.}
\begin{tabular}{cc cccc}
\toprule
$\rho$ & $\gamma$ & No Causation & $X \to Y$ & \multicolumn{2}{c}{$Y \to X$} \\
\cmidrule(lr){5-6}
 & & & & Location-shift & Non-location-shift \\
\midrule
0.6 & 0.3 & \textbf{94.7\% {\footnotesize(93.3--96.0\%)}} & \textbf{92.8\% {\footnotesize(91.2--94.4\%)}} & \textbf{90.7\% {\footnotesize(88.9--92.5\%)}} & 72.4\% {\footnotesize(69.6--75.1\%)} \\
0.6 & 0.5 & \textbf{94.7\% {\footnotesize(93.3--96.0\%)}} & \textbf{95.5\% {\footnotesize(94.2--96.7\%)}} & 78.9\% {\footnotesize(76.3--81.4\%)} & 70.9\% {\footnotesize(68.1--73.7\%)} \\
0.6 & 0.7 & \textbf{94.7\% {\footnotesize(93.3--96.0\%)}} & \textbf{96.9\% {\footnotesize(95.8--97.9\%)}} & 59.7\% {\footnotesize(56.7--62.7\%)} & 70.3\% {\footnotesize(67.4--73.1\%)} \\
0.7 & 0.3 & \textbf{94.7\% {\footnotesize(93.3--96.0\%)}} & \textbf{92.2\% {\footnotesize(90.5--93.8\%)}} & \textbf{91.7\% {\footnotesize(90.0--93.4\%)}} & 79.4\% {\footnotesize(76.9--81.9\%)} \\
0.7 & 0.5 & \textbf{94.7\% {\footnotesize(93.3--96.0\%)}} & \textbf{94.7\% {\footnotesize(93.3--96.0\%)}} & \textbf{81.3\% {\footnotesize(78.9--83.7\%)}} & 78.1\% {\footnotesize(75.5--80.6\%)} \\
0.7 & 0.7 & \textbf{94.7\% {\footnotesize(93.3--96.0\%)}} & \textbf{96.1\% {\footnotesize(94.9--97.3\%)}} & 65.2\% {\footnotesize(62.2--68.1\%)} & 77.6\% {\footnotesize(75.0--80.2\%)} \\
0.8 & 0.3 & \textbf{94.7\% {\footnotesize(93.3--96.0\%)}} & \textbf{89.4\% {\footnotesize(87.5--91.3\%)}} & \textbf{93.4\% {\footnotesize(91.8--94.9\%)}} & \textbf{86.2\% {\footnotesize(84.0--88.3\%)}} \\
${}^\dagger$0.8 & 0.5 & \textbf{94.7\% {\footnotesize(93.3--96.0\%)}} & \textbf{91.8\% {\footnotesize(90.1--93.5\%)}} & \textbf{84.9\% {\footnotesize(82.6--87.1\%)}} & \textbf{85.4\% {\footnotesize(83.2--87.6\%)}} \\
0.8 & 0.7 & \textbf{94.7\% {\footnotesize(93.3--96.0\%)}} & \textbf{93.2\% {\footnotesize(91.6--94.7\%)}} & 71.5\% {\footnotesize(68.7--74.3\%)} & \textbf{85.1\% {\footnotesize(82.9--87.3\%)}} \\
0.9 & 0.3 & \textbf{94.7\% {\footnotesize(93.3--96.0\%)}} & \textbf{80.8\% {\footnotesize(78.3--83.2\%)}} & \textbf{95.3\% {\footnotesize(93.9--96.6\%)}} & \textbf{92.0\% {\footnotesize(90.3--93.6\%)}} \\
0.9 & 0.5 & \textbf{94.7\% {\footnotesize(93.3--96.0\%)}} & \textbf{83.0\% {\footnotesize(80.6--85.3\%)}} & \textbf{89.4\% {\footnotesize(87.5--91.3\%)}} & \textbf{91.7\% {\footnotesize(90.0--93.4\%)}} \\
0.9 & 0.7 & \textbf{94.7\% {\footnotesize(93.3--96.0\%)}} & \textbf{84.4\% {\footnotesize(82.1--86.6\%)}} & \textbf{81.0\% {\footnotesize(78.5--83.4\%)}} & \textbf{91.4\% {\footnotesize(89.6--93.1\%)}} \\
\bottomrule
\end{tabular}
\label{tab:sensitivity_hyperparam}
\end{table*}

\subsection{Sensitivity to number of observations}
\label{sec:sensitivity_samplesize}

Table~\ref{tab:sensitivity_samplesize} reports the accuracy of DRCD with the default hyperparameters ($\rho=0.8$, $\gamma=0.5$) across different numbers of observations $n \in \{200, 500, 1000, 2000\}$.

Performance improves consistently with the number of observations. At $n=200$, the accuracy for $X \to Y$ is $58.0\%$ and for $Y \to X$ (non-location-shift) is $67.8\%$, reflecting the difficulty of reliable density ratio estimation with limited data. At $n=500$, all four scenarios exceed $80\%$. Performance continues to improve at $n=1000$ and $n=2000$, with the latter achieving above $85\%$ in all scenarios. These results suggest that $n \geq 500$ is a practical minimum for reliable application of DRCD.

\begin{table*}[h]
\centering
\caption{Accuracy (\%) of DRCD across different numbers of observations ($\rho{=}0.8$, $\gamma{=}0.5$). Accuracies above 80\% are shown in \textbf{bold}.}
\begin{tabular}{c cccc}
\toprule
$n$ & No Causation & $X \to Y$ & \multicolumn{2}{c}{$Y \to X$} \\
\cmidrule(lr){4-5}
 & & & Location-shift & Non-location-shift \\
\midrule
200 & \textbf{96.6\% {\footnotesize(95.4--97.7\%)}} & 58.0\% {\footnotesize(54.9--61.1\%)} & \textbf{84.3\% {\footnotesize(82.0--86.5\%)}} & 67.8\% {\footnotesize(64.9--70.7\%)} \\
500 & \textbf{95.5\% {\footnotesize(94.2--96.7\%)}} & \textbf{84.7\% {\footnotesize(82.4--86.9\%)}} & \textbf{82.2\% {\footnotesize(79.8--84.5\%)}} & \textbf{80.7\% {\footnotesize(78.2--83.1\%)}} \\
1000 & \textbf{94.7\% {\footnotesize(93.3--96.0\%)}} & \textbf{91.8\% {\footnotesize(90.1--93.5\%)}} & \textbf{84.9\% {\footnotesize(82.6--87.1\%)}} & \textbf{85.4\% {\footnotesize(83.2--87.6\%)}} \\
2000 & \textbf{95.7\% {\footnotesize(94.4--96.9\%)}} & \textbf{93.2\% {\footnotesize(91.6--94.7\%)}} & \textbf{86.5\% {\footnotesize(84.3--88.6\%)}} & \textbf{88.7\% {\footnotesize(86.7--90.6\%)}} \\
\bottomrule
\end{tabular}
\label{tab:sensitivity_samplesize}
\end{table*}

\end{document}